%% file: main.tex
\documentclass[sigconf, nonacm, pdfa]{acmart}
\newcommand\vldbdoi{10.14778/3748191.3748194}
\newcommand\vldbpages{3269 - 3283}
\newcommand\vldbvolume{18}
\newcommand\vldbissue{10}
\newcommand\vldbyear{2025}
\newcommand\vldbauthors{\authors}
\newcommand\vldbtitle{\shorttitle} 
\newcommand\vldbavailabilityurl{https://github.com/gzy02/LEGO-GraphRAG}
\newcommand\vldbpagestyle{empty} 

\usepackage[T1]{fontenc}
\usepackage{aecompl}
\usepackage{hyperref}
\usepackage{tikz}
\usepackage{algorithm}
\usepackage{algorithmic}
\usepackage{tablefootnote}
\usepackage{ulem}
\usepackage{url}
\usepackage{booktabs}
\usepackage{enumitem}
\setlist[itemize]{topsep=0pt, partopsep=0pt, parsep=0pt, itemsep=0pt}
\usepackage{multirow}
\usepackage{amsmath}
\usepackage{cleveref}
\usepackage{tcolorbox} 
\tcbuselibrary{breakable}
\usepackage{graphicx}
\usepackage{makecell}
\usepackage{tabularx}
\usepackage{color}
\usepackage{subcaption}
\usepackage{changepage}
\usepackage{soul}
\usepackage{booktabs}
\usepackage{threeparttable}
\usepackage{caption}

\usepackage[a-2b]{pdfx}

\newcommand{\ourmethod}{{LEGO-GraphRAG}}
\newcommand{\InstanceNum}{15}

\newcommand{\preRetrieval}{\text{subgraph-extraction}}
\newcommand{\retrieval}{path-retrieval}
\newcommand{\PreRetrieval}{\text{Subgraph-Extraction}}
\newcommand{\Retrieval}{Path-Retrieval}
\newcommand{\ShortPreRetrieval}{SE}
\newcommand{\ShortRetrieval}{PR}

\newcommand{\GetShortestPaths}{Shortest Path-Retrieval}
\newcommand{\GetAllPaths}{Enumerated  Path-Retrieval}
\newcommand{\getShortestPaths}{shortest path-Retrieval}
\newcommand{\getAllPaths}{enumerated  path-Retrieval}

\newcommand{\ShortRerankModel}{Reranker}
\newcommand{\llama}[2]{{Llama#1-#2B}}

\newcommand{\qwen}[2]{{Qwen#1-#2B}}

\newtheorem{definition}{Definition}
\newtheorem{example}{Example}

\newcommand{\PPRxSPR}{1}

\newcommand{\EEMxEEMxOSAR}{2}
\newcommand{\EEMxLLMxOSAR}{3}
\newcommand{\LLMxEEMxOSAR}{4}
\newcommand{\LLMxLLMxOSAR}{5}

\newcommand{\EEMxEEMxISAR}{6}
\newcommand{\EEMxLLMxISAR}{7}
\newcommand{\LLMxEEMxISAR}{8}
\newcommand{\LLMxLLMxISAR}{9}

\newcommand{\EEMxSPR}{10}
\newcommand{\LLMxSPR}{11}

\newcommand{\PPRxEEMxOSAR}{12}
\newcommand{\PPRxLLMxOSAR}{13}

\newcommand{\PPRxEEMxISAR}{14}
\newcommand{\PPRxLLMxISAR}{15}

\title{LEGO-GraphRAG: Modularizing Graph-based Retrieval-Augmented Generation for Design Space Exploration}

\author{Yukun Cao}
\affiliation{
    \institution{University of Science and Technology of China}
}
\email{ykcho@mail.ustc.edu.cn}

\author{Zengyi Gao}
\affiliation{
    \institution{University of Science and Technology of China}
}
\email{gzy02@mail.ustc.edu.cn}

\author{Zhiyang Li}
\affiliation{
    \institution{University of Science and Technology of China} 
}
\email{lizhiyang215@gmail.com}

\author{Xike Xie}
\affiliation{%
    \institution{University of Science and Technology of China}
}
\email{xkxie@ustc.edu.cn}

\author{S. Kevin Zhou}
\affiliation{%
    \institution{University of Science and Technology of China}
}
\email{s.kevin.zhou@gmail.com}

\author{Jianliang Xu}
\affiliation{%
    \institution{Hong Kong Baptist University}
}
\email{xujl@comp.hkbu.edu.hk}

\begin{document}

 \begin{abstract}
    \input{section/abstract.tex}

 \end{abstract}

 \maketitle

\pagestyle{\vldbpagestyle}
\begingroup\small\noindent\raggedright\textbf{PVLDB Reference Format:}\\
\vldbauthors. \vldbtitle. PVLDB, \vldbvolume(\vldbissue): \vldbpages, \vldbyear.\\
\href{https://doi.org/\vldbdoi}{doi:\vldbdoi}
\endgroup
\begingroup
\renewcommand\thefootnote{}\footnote{\noindent
This work is licensed under the Creative Commons BY-NC-ND 4.0 International License. Visit \url{https://creativecommons.org/licenses/by-nc-nd/4.0/} to view a copy of this license. For any use beyond those covered by this license, obtain permission by emailing \href{mailto:info@vldb.org}{info@vldb.org}. Copyright is held by the owner/author(s). Publication rights licensed to the VLDB Endowment. \\
\raggedright Proceedings of the VLDB Endowment, Vol. \vldbvolume, No. \vldbissue\ %
ISSN 2150-8097. \\
\href{https://doi.org/\vldbdoi}{doi:\vldbdoi} \\
}\addtocounter{footnote}{-1}\endgroup

\ifdefempty{\vldbavailabilityurl}{}{
\vspace{.3cm}
\begingroup\small\noindent\raggedright\textbf{PVLDB Artifact Availability:}\\
The source code, data, and/or other artifacts have been made available at \url{\vldbavailabilityurl}.
\endgroup
}


 \input{section/intro.tex}

 \input{section/pre_new.tex}

\input{section/relatedworks.tex}\input{section/method.tex}

\input{section/instance.tex}\input{section/experiment.tex}

\input{section/results.tex}

\input{section/discussion.tex}

\input{section/conclusion.tex}

\input{section/acknowledgement.tex}

\newpage

\appendix
\setcounter{section}{1} 
\renewcommand{\thesection}{\Alph{section}} 
\bibliographystyle{ACM-Reference-Format}
\normalem
\bibliography{bib/FRAG.bib, bib/GRAG.bib, bib/LEGO.bib}

\end{document}

%% file: section/abstract.tex
GraphRAG integrates (knowledge) graphs with large language models (LLMs) to improve reasoning accuracy and contextual relevance. 
Despite its promising applications and strong relevance to multiple research communities, such as databases and natural language processing, GraphRAG currently lacks modular workflow analysis, systematic solution frameworks, and insightful empirical studies. 
To bridge these gaps, we propose {\bf LEGO-GraphRAG}, a modular framework that enables: {\bf 1)} fine-grained decomposition of the GraphRAG workflow, {\bf 2)} systematic classification of existing techniques and implemented GraphRAG instances, and {\bf 3)} creation of new GraphRAG instances. 
Our framework facilitates comprehensive empirical studies of GraphRAG on large-scale real-world graphs and diverse query sets, revealing insights into balancing reasoning quality, runtime efficiency, and token or GPU cost, that are essential for building advanced GraphRAG systems.

%% file: section/intro.tex
\section{Introduction}
\label{section: introduction}

Recent advancements in large language models (LLMs) have highlighted their strengths in semantic understanding and contextual reasoning, enabled by extensive pre-training on vast corpora. Despite these strengths, LLMs often struggle with domain-specific queries and complex contexts, frequently generating ``hallucinations'', in which outputs appear credible but lack factual accuracy. 
Retrieval-augmented generation (RAG) addresses these limitations by integrating external knowledge to enhance factual accuracy and contextual relevance.
Early RAG implementations, however, often relied on document retrieval methods that introduced noise and excessive context~\cite{lewis2021retrieval,ColBERT,yu2023generate, HyDE}. 
This limitation has led to a growing focus on graph-based RAG systems, known as {\it GraphRAG}~\cite{GraphRAG,surveyTJFD,GRAG,SubgraphRAG}. 

In GraphRAG, conventional document retrieval is replaced by graph-based retrieval, leveraging the structured and relational nature of graphs to extract query-specific reasoning paths that provide more precise and contextually relevant support for LLM reasoning.
Typically, the GraphRAG workflow consists of two primary phases:
\begin{itemize}[leftmargin=12pt]
\item {\bf Retrieval:} This phase retrieves knowledge from graphs by identifying reasoning paths relevant to the query. 
\item {\bf Augmented generation:} The retrieved reasoning paths are used to augment the LLM prompt, enhancing its ability in reasoning and generating accurate and contextually relevant outputs. 
\end{itemize}
Existing GraphRAG studies~\cite{GraphRAG,GCR, RoG,LessIsMore,StructGPT,ref:kelp,ToG,DoG}
have demonstrated the potential of integrating graph data management techniques with GraphRAG. However, despite their promise, these studies are still in their early stages and face two significant challenges. First, they lack foundational support in addressing the scalability of graph algorithms, which is crucial for managing large-scale graphs and reducing query latency in real-world applications. Second, there is a knowledge gap regarding the impact of semantic information on graphs on the performance of GraphRAG. Specifically, it is unclear which strategies are most effective for leveraging semantic information from both the query and the graph, and which type of semantic model is best suited for different query scenarios. Resolving these uncertainties is essential to improving the quality, efficiency, and cost of GraphRAG across various query scenarios.




\textbf{Our Motivation.}
Building on the strengths of database research, we aim to address key challenges in GraphRAG and lay the groundwork for future advancements. 
While efforts like Modular RAG~\cite{modularRag} have modularized general RAG, they fall short of addressing the GraphRAG's workflows and needs.
We identify three critical gaps:

\begin{itemize}  [leftmargin=12pt]
\item 
\textbf{Need for a Unified Framework for Categorizing and Analyzing GraphRAG Solutions.}
GraphRAG solutions consist of diverse technologies, including algorithmic approaches (e.g., random walk, PageRank) and neural network-based methods (e.g., end-to-end and LLM models), each playing a distinct role in their functionality. The lack of a unified categorization for systematically analyzing these techniques—despite prior efforts such as Modular RAG  ~\cite{modularRag} and other surveys ~\cite{RAGsurvey1,RAGsurvey2,RAGsurvey3,RAGsurvey4} on LLMs with graphs—hinders effective research summarization and the identification of promising candidates for further study.

\item 
\textbf{Need for Modular Retrieval in GraphRAG.}
Current implementations of GraphRAG often treat the retrieval phase as a single, monolithic process, making it difficult to isolate, analyze, and optimize individual components. A modular approach is desirable to facilitate the development of more efficient and effective retrieval mechanisms.
Moreover, while Modular RAG~\cite{modularRag} modularizes general RAG workflows, it neither dissects the core retrieval process of GraphRAG nor clarifies its distinction from general RAG.

\item 
\textbf{Absence of a GraphRAG Testbed for New Instance Design and Evaluation.}
Optimizing GraphRAG performance requires navigating a complex trade-off between runtime efficiency and reasoning accuracy, influenced by various design factors. A GraphRAG testbed is needed to generate diverse implementation instances, allowing users to evaluate, compare, and apply different approaches while providing clear guidelines and trade-offs for constructing optimal GraphRAG instances tailored to specific scenarios.
However, Modular RAG  ~\cite{modularRag} and related surveys ~\cite{RAGsurvey1,RAGsurvey2,RAGsurvey3,RAGsurvey4} focus on conceptual overviews and lack implementation or evaluation support for GraphRAG.
\end{itemize}

\textbf{Our Contributions.}
We presents the first empirical study on GraphRAG and introduces {\it {\ourmethod}}, a unified and modular framework that provides insights to guide future research through contributions in both framework design and empirical analysis. 

For Framework Design:

\begin{itemize} [leftmargin=12pt]
    \item We propose LEGO-GraphRAG, a modular framework dividing the retrieval phase into two flexible modules: {\it {\preRetrieval}} and {\it {\retrieval}}, and classifies the techniques into {\it structure-based} and {\it semantic-augmented} methods for  each module.
    \item LEGO-GraphRAG’s modular framework, with categorized techniques, supports implementing all existing GraphRAG instances while enabling the creation of new ones, promoting both standardization and innovation.
    \item We identify essential design factors, including reasoning quality, efficiency, and cost, providing a structured trade-off analysis to guide the development of GraphRAG instances.
\end{itemize}
For Empirical Research:

\begin{itemize}[leftmargin=12pt]
    \item We study a comprehensive set of GraphRAG instances, including $7$ existing implementations and $16$ new instances generated by LEGO-GraphRAG. These instances were extensively evaluated on large-scale real graphs (i.e., Freebase) and $5$ commonly used GraphRAG query sets, covering various query scenarios.
    \item We suggest several modular configurations and a number of improvements to existing implementations for improved reasoning quality and balancing efficiency and cost.
\end{itemize}





\subsection{Related Works}
\label{section: rw}

\subsubsection{RAG and GraphRAG}
Early text-based RAG systems rely on basic retrieval techniques, such as text chunking and cosine similarity for ranking~\cite{lewis2021retrieval}, which are prone to retrieving noisy and irrelevant information, leading to lower-quality results~\cite{HyDE}. To address this, recent works have 
introduced both pre- and post-retrieval improvements. 
Pre-retrieval methods~\cite{StepBackRAG} enhance the input query, 
while post-retrieval techniques~\cite{G-RAG, R4} refine the ranking and filtering of retrieved results. 
Toolkits like LlamaIndex and LangChain~\cite{LlamaIndex, LangChain} offer modular control over these stages, improving overall retrieval precision and system interpretability~\cite{HyDE}.
However, refining the retrieval stage to reliably suppress irrelevant content remains a key challenge~\cite{GraphRAG}. 

GraphRAG has recently emerged as a promising direction by representing knowledge in graph form, enabling more structured retrieval, multi-hop reasoning with  better interpretability~\cite{GraphRAG, RAGvsGraphRAG, 2501.13958}.
Compared to unstructured text retrieval, graph-based retrieval offers reduced noise, better coverage of entity relations,
and lowered token overhead during inference~\cite{ToG2.0, GraphRAG, NewArticles, Podcast}.\footnote{A detailed comparison between GraphRAG and text-based RAG is provided in our technical report B.10.} 
Existing instances (or implementations) in GraphRAG have drawn heavily from knowledge-base question answering (KBQA) techniques, utilizing both information retrieval methods~\cite{ref:wikimovie, EmbedKGQA} and semantic parsing models~\cite{QGG, HGNet} to identify relevant subgraphs or reasoning paths. 
Microsoft GraphRAG~\cite{GraphRAG} was an early effort exploring the advantages of graph-based over text-based retrieval, focusing on graph construction from text and precomputation. 
GCR~\cite{GCR} combines LLMs and beam search to refine reasoning paths, improving retrieval alignment with queries. RoG~\cite{RoG} and GSR~\cite{LessIsMore} employ Personalized PageRank (PPR) to retrieve query-specific subgraphs, while KELP~\cite{ref:kelp} refines reasoning paths using fine-tuned models like BERT~\cite{bert}. These studies highlight the diverse graph-based and neural network-based techniques employed in GraphRAG, showcasing the extensive potential of graphs in enhancing the reasoning quality of LLMs.

\subsubsection{LLMs with (Knowledge) Graphs}
GraphRAG aligns with the broader research to integrate LLMs with structured knowledge, such as KGs. Surveys~\cite{LLM_KG_Roadmap, LLM_KG_Trend} outline three paradigms: KG-enhanced LLMs, LLM-augmented KGs, and their joint integration.
Moreover, \cite{SurveyKPLM} reviews knowledge-enhanced pre-trained language models (e.g., LLMs) for language understanding and generation, and \cite{LLM_interface} examines LLMs as interfaces for data pipelines, highlighting their integration with KGs in AI systems.
Our work fits within the KG-enhanced LLMs paradigm and aims to contribute a modular GraphRAG framework to support systematic design and evaluation of reasoning-augmented LLMs with graphs.

%% file: section/pre_new.tex
\section{Preliminaries}
\label{section: GraphRAG}

\subsection{Problem Formalization}

GraphRAG is designed to integrate structured knowledge into the reasoning process of LLMs, enhancing their ability to generate content that is both accurate and contextually relevant.
In GraphRAG, knowledge is typically represented in the form of Text-Attributed Graphs (TAGs), where both nodes and edges are enriched with textual information, with knowledge graphs serving as a typical example.
For clarity and without loss of generality, the graphs referred to in this paper will follow the structure outlined in Definition~\ref{def:graph}.

\begin{definition}[{\bf Graph in GraphRAG}]
	\label{def:graph}
In GraphRAG, a graph is defined as a directed labeled graph \( G = (V, E) \), where \( V \) represents the set of nodes (entities), and \( E \) denotes the set of directed edges that signify relations between those entities. Each node \( v \in V \) and each edge \( e \in E \) carry semantic information. Specifically, nodes represent entities with associated semantic attributes (e.g., descriptions), while edges represent relations with semantic contexts (e.g., relation types). 

Formally, the graph in GraphRAG is represented as a collection of triples: $G = \{ \tau_i = (s_i, r_i, t_i) \mid s_i, t_i \in V, r_i \in E \}$ , 
where each \( \tau_i = (s_i, r_i, t_i) \) is a triple representing that the source entity \( s_i \) is connected to the target entity \( t_i \) by the relation \( r_i \), encapsulating structured, domain-specific knowledge. While a single triple \( \tau_i \) is a basic unit of knowledge, a sequence of connected triples forms reasoning paths that support more advanced inferences.
\end{definition}

In practice, graphs are built from domain knowledge
or text (e.g., Microsoft GraphRAG~\cite{GraphRAG} with auxiliary summaries; see Definition~\ref{def:graph_con}). As text-to-graph construction is well-studied~\cite{GraphRAG,LightRAG,KAG,ref:hipporag,sac-kg} and not central to our framework, we treat it as optional but include a GraphRAG instance with textual information in our analysis (Section~\ref{exp:instance} e).

\begin{definition}[{\bf Graph Construction ({\it Optional})}]
\label{def:graph_con}
In GraphRAG, graph construction is an optional step that builds a text-attributed graph from large-scale text,  and may also precompute subgraphs and summaries to enhance GraphRAG.
Specifically:
\begin{itemize}[leftmargin=10pt]
    \item \textbf{Construction:} Identify entities and relations from text and represent them as nodes and edges in the graph.

    \item \textbf{Precomputation ({\it Optional}):} Compute subgraphs and relevant textual summaries from the constructed graph, such as via graph clustering or community detection. 
\end{itemize}
\end{definition}

With text-attributed auxiliary graphs, GraphRAG operates in two main phases: \textit{retrieval} and \textit{augmented generation} phases. During the retrieval phase (Definition~\ref{def:retrieval}), the system extracts relevant entities and reasoning paths from the graph, aligning them with the query context.
In the augmented generation phase (Definition~\ref{def:generation}), these retrieved reasoning paths are incorporated to enhance LLM's content generation capabilities.

\begin{definition}[{\bf  Retrieval Phase}]
	\label{def:retrieval}
This phase starts by processing the query $q$ to extract relevant entities and relations that correspond to nodes and edges in the graph $G$: $\textit{Extract} (q, G) \rightarrow \epsilon_q=(\{v_i^{(q)}\}, \{e_i^{(q)}\})$, 
where $\epsilon_q$ is the set of entities and relations extracted.\footnote{ Extracting entities/relations from the semantic context of $q$ has been well-studied~\cite{NSM, EmbedKGQA, StructGPT} and is beyond the scope of this paper.
Following prior works on reasoning tasks in graphs~\cite{UHop,berant-etal-2013-semantic,yih-etal-2014-semantic,complex2simple,EmbedKGQA}, queries are typically categorized by the minimum number of reasoning hops required to reach the answer from the query entities or relations (e.g., one-hop queries or multi-hop queries). 
In addition, queries can also be categorized by the number of answers and the number of query entities involved. 
}

Using the extracted entities $\{v_i^{(q)}\}$ and relations $\{e_i^{(q)}\}$ in $\epsilon_q$, a variety of methods (e.g., search algorithms, neural network-based retrieval/ranking models) are applied to retrieve reasoning paths $ \mathcal{P}_q$ 
that connect $\epsilon_q$ to potential target answers, potentially spanning multiple hops.
The retrieval process is thus formalized by: $\textit{Retrieve} (\epsilon_q, G) \rightarrow \mathcal{P}_q=\{P_i^{(q)}\}$
where each reasoning path $P_i^{(q)}\in \mathcal{P}_q$ of length $k$ is represented as a sequence of triples $\langle \tau_1, \tau_2, \ldots, \tau_k \rangle$: 
\begin{align*}
\small
P_i^{(q)} = \langle \tau_1, \tau_2, \ldots, \tau_k \rangle=\langle (s_1, r_1, t_1), (t_1, r_2, t_2), \ldots, (t_{k-1}, r_k, t_k) \rangle 
\end{align*}

\end{definition}

\begin{definition}[{\bf Augmented Generation Phase}]
	\label{def:generation}
In this phase, the retrieved reasoning paths $\mathcal{P}_q$ are merged with the original query $q$, forming an augmented prompt $q'$: $\textit{Augment}(q, \mathcal{P}_{q}) \rightarrow q'$.
This augmented prompt is then input to the LLMs, enhancing their ability to generate more accurate and contextually relevant content for the final answer: $	\textit{Generate}(q', \text{LLM}) \rightarrow \textit{final answer}$.
\end{definition}

\begin{figure}[t]
    \includegraphics[width=0.98\linewidth]{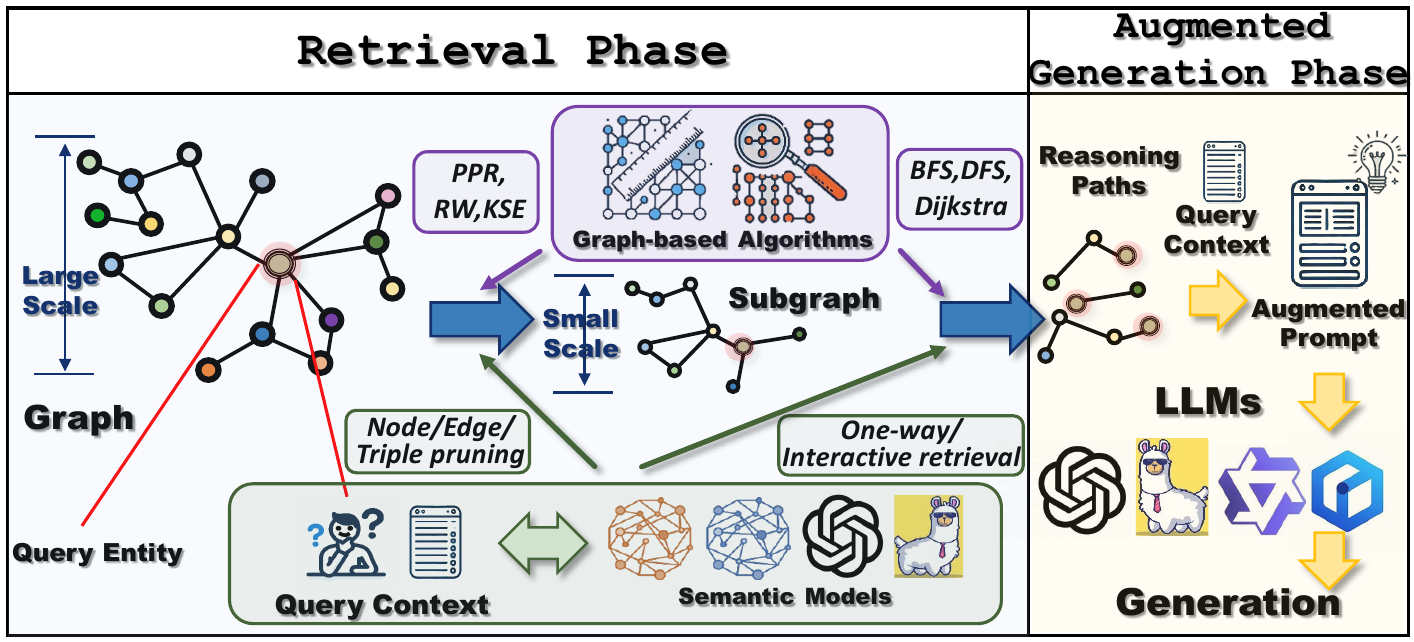}
    \vspace{-5pt}
    \caption{The GraphRAG Workflow} 
    \label{GraphRAG}
\end{figure}

Figure~\ref{GraphRAG} illustrates the GraphRAG workflow, where retrieval forms the foundation by supplying key reasoning paths for LLM augmentation and generation. As these stages hinge on retrieval quality and LLM strength, this work focuses on analyzing the modular design and solution space of the retrieval phase.


%% file: section/method.tex
\section{LEGO-GraphRAG Framework}
\label{section: Modular GraphRAG}


\subsection{Overview}
Table~\ref{tab:design-space} outlines the design framework of LEGO-GraphRAG, structured along two key dimensions: {\bf modules} and {\bf method types}. Specifically, the core retrieval phase is divided into two modules: the \textit{subgraph-extraction} module and the \textit{path-retrieval} module. Each module incorporates two solution types: \textit{structure-based} methods and \textit{semantic-augmented} methods.
Accordingly, GraphRAG instan\\ces within the LEGO-GraphRAG framework are classified into five distinct groups based on the combinations of modules and solution types used.  
Table~\ref{tab:flat_table} provides a detailed breakdown of these groups, encompassing both existing methods and new instances developed in our empirical study.
In general, the LEGO-GraphRAG framework is materialized by two key considerations.
\begin{itemize}[leftmargin=10pt]
    \item {\bf Natural Segmentation of the GraphRAG Process.} 
    The reasoning path retrieval process is naturally segmented into two phases:
    a) extracting a query-relevant subgraph to scale down the search space; and b) meticulously retrieving reasoning paths from the extracted subgraph.
    \item 
    {\bf Facilitation of GraphRAG Research Analysis.} 
    The framework enables a comprehensive analysis of recent GraphRAG advancements, which predominantly focus on enhancing one or both modules, as summarized in Table~\ref{tab:flat_table}. 
\end{itemize}

\vspace{-5pt}
\subsubsection{{\bf {\PreRetrieval} {\bf vs.} {\Retrieval} Modules}}


The subgraph-extraction (SE) module aims to enhance the effectiveness and efficiency of reasoning path retrieval by reducing the search space from the full graph to a smaller set of query-relevant subgraphs. The path-retrieve (PR) module then operates on these subgraphs to retrieve reasoning paths using specific methods. 
Based on the retrieval phase of GraphRAG formalized in Definition~\ref{def:retrieval}, the two modules are formalized as follows.

\begin{definition}[\textbf{\PreRetrieval ~(\ShortPreRetrieval)}]
\label{def:pf}
\vspace{-5pt}
Given a query $q$, a graph $G = (V, E)$, and a set of entities and relations $\epsilon_q$ derived from specified query $q$, the SE module aims to extract a query-specific subgraph $g_q \subseteq G$, defined as:~~~~
$\ShortPreRetrieval(G, q, \epsilon_q) \rightarrow g_q$.
\end{definition}

\begin{definition}[\textbf{\Retrieval~(\ShortRetrieval)}]
\vspace{-5pt}
Given an extracted subgraph $g_q \subseteq G$ and the entity and relation set $\epsilon_q$ derived from query $q$, the process of this module obtains a reasoning path set $\mathcal{P}_q = \{P_i^{(q)}\}$ with the following process:~~~~ $\ShortRetrieval(G\text{~or~} g_q, q, \epsilon_q)  \rightarrow \mathcal{P}_q = \{P_i^{(q)}\}$.
\end{definition}

\input{tables/solutions.tex}
\input{tables/pipelines.tex}

\vspace{-5pt}
Of the two modules, the {\it \retrieval} module is essential for any GraphRAG instance, while the {\it \preRetrieval} module is optional.
The optionality of the {\it \preRetrieval} module depends largely on the graph size, primarily due to computational efficiency. 
For smaller graphs, where the node and edge space remain manageable, direct retrieval of reasoning paths is feasible, making the {\it \preRetrieval} module optional. 
However, for large-scale graphs, the exponential growth in the number of nodes, edges, and paths significantly complicates the retrieval {\cite{BigGraphVLDB,PrismX,SummaryBigGraph,SysBigGraph}}. 
The {\it \preRetrieval} module addresses this by extracting a query-relevant subgraph, reducing the retrieval space and improving computational efficiency and relevance.
After passing through these modules, the refined reasoning paths enhance the reasoning abilities of LLMs during the generation phase, as detailed in Definition~\ref{def:generation}.

\vspace{-5pt}
\subsubsection{\bf Structure-based {\bf vs.} Semantic-augmented Methods}

For each module, systematically classifying and summarizing potential solutions is important, as it enables the identification of potential strategies and provides a deeper understanding of existing implementations.  
At their core, the workflows of the {\it \preRetrieval} and {\it \retrieval} modules share potential similarities, as both involve 
retrieving a small-scale but query-relevant subset from the graph.
Accordingly, the design solutions for these modules can be interwoven and further categorized into two complementary categories: {\it structure-based methods}, focusing on the graph's topological properties, and {\it semantic-augmented methods}, utilizing the semantic information of both the graph and the query, as detailed in Table~\ref{tab:design-space}.

Structure-based methods (SBE and SBR) use graph algorithms~\cite{Korf1985DepthFirstIA,ThunderRW,FastRWR,FastPPR,Topppr,PPRSurvey} to iteratively explore nodes (entities) and edges (relations), 
constructing subgraphs or reasoning paths.
Semantic-augmented methods (SAE, OSAR, and ISAR) exploit the semantic relevance between the query and the nodes, edges, and triples of graphs, using two types of semantic models: end-to-end models (EEMs) {\cite{reimers-2019-sentence-bert,reimers-2020-multilingual-sentence-bert,thakur-2020-AugSBERT}} and large language models (LLMs) {\cite{LLMSurvey,ToG,GraphInsight}}. 
The two types of models are applied differently. 
SAE/OSAR methods refine candidate subgraphs/paths for better semantic alignment.
ISAR methods incorporate semantic evaluation directly into the path-retrieval process, using EEMs/LLMs to interactively identify relevant reasoning paths.
In the sequel, we investigate the design solutions of the two modules, subgraph-extraction (Section~\ref{subsection: design solutions of preretrieval}) and path-retrieval (Section~\ref{subsection: design solutions of retrieval}), with the aforementioned two types of methods, structure-based and semantic-augmented methods.

\vspace{-5pt}
\subsection{Design Solutions of {\PreRetrieval}}
\label{subsection: design solutions of preretrieval}

The design solution of the subgraph-extraction module is comprised of \textit{structure-based extraction} (\textbf{SBE} in Section~\ref{subsec:sbe}) and \textit{semantic-based extraction} (\textbf{SAE} in Section~\ref{subsec:sae}).

\vspace{-5pt}
\subsubsection{{\bf Structure-Based Extraction ({\bf SBE})}} 
\label{subsec:sbe}
These solutions exploit the structural information of a graph \( G = (V, E) \), starting with the query-relevant entities \( \{v_i^{(q)}\} \in \epsilon_q \),\footnote{While subgraphs can be built from relationship \( e_i^{(q)} \in \epsilon_q \), practical methods predominantly use entities (nodes) as the basis for subgraph construction~\cite{ref:kelp,ToG,ChatKBQA,GCR,LessIsMore}.} to identify corresponding nodes (entities) and edges (relations) that are pertinent to the query \( q \). This process constructs smaller subgraphs centered around each \( v_i^{(q)} \), which are then aggregated to produce the query-related subgraph \( g_q \). Formally, the procedure is defined as:
\begin{equation*}
\small
\ShortPreRetrieval(G, q, \epsilon_q) \rightarrow g_q = \bigcup\nolimits_{v_i^{(q)} \in \epsilon_q} g_{(q,v_i^{(q)})}
\vspace{-5pt}
\end{equation*}

Here, \( g_q \) represents the union of the subgraphs \(g_{(q,v_i^{(q)})} \), each focusing on an entity \( v_i^{(q)} \) in \( \epsilon_q \). For each query entity, three extraction strategies can be applied: {\it random walk-based}, {\it neighborhood-based}, and {\it structural importance-based} strategies.

{\it \underline{a) Random Walk-based.}} The {\it random walk (RW)} algorithm {\cite{RandomWalk,kRW}} and its variants {\cite{LRW,RWR,FastRWR,SecondRW,SpaceRW} enumerate all entities in $\epsilon_q$, and, for each entity $v_i^{(q)}$,
iteratively expand a subgraph by randomly selecting edges and nodes at each step. This simple yet effective strategy gradually constructs a subgraph of the desired size, making it a commonly used baseline method~\cite{NSM,RoG,LessIsMore,GCR}.

{\it \underline{b) Neighborhood-based.}} This strategy aims to capture the local structure around an entity \( v_i^{(q)}\in \epsilon_q \) by expanding its neighborhood. The most representative method is {\it K-hop subgraph-extraction (KSE)} {\cite{NSM}}, where the subgraph is constructed by including all nodes \( v_j \) and edges \( e_{ij} \) such that the shortest path distance \( d(v_i^{(q)}, v_j) \leq K \). 
This approach ensures that the subgraph captures the immediate relational context surrounding the query entity, progressively including nodes and edges within the specified radius (i.e., $K$-hops).

{\it \underline{c) Structural Importance-based.}} These methods  {\cite{PageRank,PPR,PPRSurvey}} identify nodes based on their structural importance relative to an entity \( v_i^{(q)} \). Nodes are ranked by importance scores, and a subset of high-ranking $N_{ppr}$ nodes, along with their directly connected edges, are selected to form the subgraph.
Prominent methods, such as {\it PageRank} {\cite{PageRank}} and {\it Personalized PageRank (PPR)} {\cite{PPR}}, compute a relevance score for each node based on topological position, link density and centrality.
Specifically, PageRank assigns an importance score \( S(v_j) \) to each node \( v_j \), which is iteratively updated based on both its local connectivity and the scores of its neighboring nodes, thereby capturing its global significance. Personalized PageRank (PPR) enhances this process by ``personalizing'' the distribution towards the query entity \( v_i^{(q)} \), ensuring that nodes closer to \( v_i^{(q)} \) receive higher scores.
While PPR is a widely adopted approach for SBE solutions, its computational cost scales with the size of the graph, presenting a considerable bottleneck for real-time GraphRAG pipelines. Section~\ref{exp:PreRetrieval} discusses this efficiency challenge in detail.


\vspace{-5pt}
\subsubsection{{\bf Semantic-Augmented Extraction ({\bf SAE})}}
\label{subsec:sae}
Semantic models play a key role in facilitating subgraph-extraction by assessing the semantic relevance of graph components.
These models fall into two main categories: 
End-to-End Models (EEMs) and Large Language Models (LLMs).

{\it \underline{a) End-to-End Models (EEMs).}} 
EEMs process the text associated with graph components 
(i.e., nodes, edges, triples) and queries to compute semantic relevance.
They either output statistical properties and embeddings for relevance calculation or directly provide relevance scores.
Formally, for nodes, edges, or triples \( v, e, \text{~or~} \tau \in G \) and a query \( q \), the goal is: $\text{EEMs} (v \text{ or } e \text{ or } \tau, q) \to \text{scores}.$ 
Using these scores, the most relevant graph components are identified for subgraph-extraction.
EEM can be categorized into two main types: {\it Statistical Models} {\cite{TFIDF,BM25,BM25+}} and {\it Embedding Models/Re-Rankers} {\cite{reimers-2019-sentence-bert,BGE}}. The former relies on statistical computations, while the latter consists of end-to-end neural network models to produce embeddings or rank relevance.

   {\it \uwave{a).1 Statistical Models.}}
Statistical models are foundational in text analysis and semantic modeling, relying on metrics 
like term frequency (TF) and inverse document frequency (IDF)  {\cite{TFIDF}}. 
Popular methods include {\it BM25}  {\cite{BM25}} for ranking relevance in information retrieval, {\it Latent Semantic Analysis (LSA)}  {\cite{LSA}}  for uncovering latent semantic structures via dimensionality reduction, and {\it Latent Dirichlet Allocation (LDA)}  {\cite{LDA}}  for topic modeling, which aids 
semantic similarity through topic distributions.

For example, BM25 treats textual information attached to graph components (nodes, edges, or triples) as ``documents'' within a corpus (the graph). 
During precomputation, the text of each graph component is tokenized, enabling the calculation of TF and IDF. 
When a query is issued, its text undergoes tokenization, 
and BM25 computes a relevance score for each graph component based on the query terms and their frequency distributions across the graph. 
Statistical models are efficient, with low computational costs and short execution times. They are suitable for simpler semantic tasks or as baselines in semantic-augmented workflows.



{\it \uwave{a).2 Embedding Models/Re-Rankers.}} Embedding models and re-rankers use pre-trained neural network (NN) models with smaller parameter sizes (millions) compared to LLMs (billions). Trained on large corpora, they generate dense embeddings that encode rich semantic information {\cite{reimers-2019-sentence-bert}}. Embedding models (e.g., Sentence-Transformers {\cite{MiniLM}}) independently generate embeddings for queries and graph components (nodes, edges, triples). Semantic similarity is computed using metrics {\cite{jaccard, ChandrasekaranM21}} like cosine similarity: 
\begin{equation*}
\small
\text{ST}(v \text{ or } e \text{ or } \tau, q) \to \cos(\mathbf{emb}(v \text{ or } e \text{ or } \tau), \mathbf{emb}(q))
\end{equation*}
Re-Rankers (e.g., BGE-Reranker  {\cite{BGE}}) take combined query and graph components as inputs and output a similarity score, allowing for deeper contextual interactions:
\begin{equation*}
\small
\text{BGE}(v \text{ or } e \text{ or } \tau, q) \to \text{NeuralNetworks}(\mathbf{emb}(v \text{ or } e \text{ or } \tau) \oplus \mathbf{emb}(q))
\end{equation*}
These models generally require longer execution and pre-training times compared to statistical models. However, their balance of efficiency and semantic precision makes them a common choice for semantic modeling in  GraphRAG. 
Domain-specific fine-tuning can further enhance these models' performance {\cite{RoG,ref:kelp,LessIsMore,GCR}}, though it comes at an additional cost. 



{\it \uwave{a).3 SAE with EEMs.}}
In the {\preRetrieval} module, EEMs prune nodes, edges, or triples from graph $G$ based on their semantic relevance to the query, 
reducing computational overhead.
However, directly applying EEMs to large graphs is costly. 
Thus, it is desired to have a \textit{pre-filtering} step before applying semantic models, which extracts a smaller subgraph \( g \in G \) (e.g., via SBE methods-PPR) to focus on relevant components~\cite{ref:kelp,LessIsMore}. 
Thus, this subgraph pruning process generally follows three key strategies: {\it node pruning}, {\it edge pruning}, and {\it triple pruning}.


\begin{itemize}[leftmargin=10pt]
\item 
{ Node Pruning (NP):} Nodes $v \in g$ 
are ranked by semantic relevance scores to query $q$. Top $N_v$ nodes are retained, ensuring connected edges are included:  $    g_{(q, v)}= \{v \mid v \text{ in top } N_v\} \cup \{e \mid e \text{ connects } g_{(q, v)}\}$.
\item 
{Edge Pruning (EP):} Edges $e \in g$ are scored, with top $N_e$ edges retained. Disconnected nodes are removed: $g_{(q, e)} = \{e \mid e \text{ in top } N_e\} \\ \cup \{v \mid v \text{ connected by } g_{(q, e)}\}$.
\item 
{Triple Pruning (TP):} 
Top $N_\tau$ triples $\tau \in g$ are selected, forming a minimal subgraph containing these triples: 
$g_{(q, \tau)} = \{\tau \mid \tau \text{ in top } N_\tau\}$.

\end{itemize}

Batch processing can be used for EEMs to efficiently compute relevance scores for all graph components in \( g\), streamlining subgraph-extraction:
\begin{equation*}
\small
\ShortPreRetrieval(G, q, \epsilon_q), EEMs \rightarrow  
g_q= g_{(q, v/e/\tau)}
\end{equation*}

{\it \underline{b) Large Language Models (LLMs).}}
LLMs, such as Llama {\cite{llama,llama2,llama3}} , Qwen  {\cite{qwen2}}, and GPT  {\cite{ref:gpt3,ref:gpt4}}, are large pre-trained language models known for their ability to capture nuanced semantics and contextual understanding due to extensive training on large-scale corpora. Unlike EEMs, LLMs provide flexible evaluation of graph components (nodes, edges, triples) through prompt-based interactions.
For instance, to evaluate the node relevance to a query,  LLMs can generate relevance scores or directly a ranked list of entities. 
Formally, this process is represented as:
\begin{equation*}
\small
\label{eq:LLMs_score}
\text{LLMs}(v \text{ or } e \text{ or } \tau, q, \text{Prompt}) \rightarrow \text{scores} \text{ or } \text{ranked list}  
\end{equation*}    

In practice, LLM outputs often fall short of prompt requirements due to output length limits and inherent limitations (e.g., hallucination~\cite{ref:hallucination}). Generated entity lists (from subgraph-extraction module) or reasoning paths (from path-retrieval module) may be sparse or low-quality. Mitigation approaches include: prompt tuning ~\cite{lift}, fine-tuning ~\cite{longlora} for long-context understanding, or adopting more capable LLMs.  Moreover, we explore some general purpose strategies to address this challenge in Section~\ref{subsec:LLMsQ}.

While fine-tuning LLMs on domain-specific knowledge improves performance for targeted queries, it is more resource-intensive than tuning embedding models or re-rankers\footnote{Details about fine-tuning for both EEMs and LLMs are in the technical report B.1.} and may generalize poorly across scenarios~\cite{SubgraphRAG}. LLMs also have high inference costs due to the model size and autoregressive decoding. 
Hence, we recommend using them selectively and offer non-LLM alternatives in Section~\ref{exp:instance}. Cost-efficient fine-tuning methods (e.g., LoRA~\cite{lora}, QLoRA~\cite{qlora}) and inference optimizations (e.g., pruning~\cite{evop}, quantization~\cite{sustainable}, KV caching~\cite{KVLLM}, parallel decoding~\cite{hardware}, and semantic compression~\cite{chunkkv}) further reduce overhead.

\uwave{{\it b).1 SAE with LLMs.}}
The subgraph extraction process using LLMs mirrors that of EEMs and can be formalized as follows:
\begin{equation*}
\small
\ShortPreRetrieval(G, q, \epsilon_q), LLMs, Prompts \rightarrow  g_q= g_{(q, v/e/\tau)}
\end{equation*}
Even on subgraph \( g \), LLM-based semantic evaluation remains costly and faces token constraints, making it hard to process $g$ with a single LLM call. To mitigate this, an additional \textit{pre-filtering} step is applied, where candidate components within $g$ can be ranked and pruned via lightweight heuristics (e.g., random selection or EEM-based similarity selection) before invoking LLMs.
Subsequently, the refined $g$ can be passed to LLMs for evaluation within a single call.\footnote{
Multiple LLM calls can be used to evaluate all graph components in \( g \), but this incurs substantial overhead during subgraph-extraction and is typically avoided.}


\uwave{{\it b).2 Other Methods with LLMs and Limitations.}}
In certain scenarios, LLMs fine-tuned on domain knowledge can act as agents {\cite{RoG,ChatKBQA}} to generate extraction rules (e.g., SPARQL queries~\cite{ScalableSPARQL, SurveySPARQL, ChatKBQA}).
For example, in response to a query, RoG~\cite{RoG} utilizes LLMs fine-tuned on domain-specific knowledge graphs to identify key relationships between entities, which are then used to construct SPARQL queries that extract relevant triples, forming the query-relevant subgraph.
The subgraph containing these triples ultimately serves as the query-relevant subgraph. While LLMs offer direct subgraph-extraction without pruning, these approaches are limited by high computational costs, reliance on accurate templates, and reduced generalizability for diverse queries. Fine-tuning LLMs for domain-specific tasks further adds to the resource burden, making this approach less scalable for subgraph-extraction on large-scale graphs.



\subsection{Design Solutions of {\Retrieval}}
\label{subsection: design solutions of retrieval}

The design solution of the path-retrieval module is comprised of {\it structure-based retrieval} ({\bf SBR} in Section~\ref{subsubsec:SBR}), {\it one-way semantic-augmented retrieval} ({\bf OSAR} in Section~\ref{subsubsec:OSAR}), and {\it interactive semantic augmented retrieval} ({\bf ISAR} in Section~\ref{subsubsec:ISAR}).

\subsubsection{{\bf Structure-Based Retrieval ({\bf SBR})}} 
\label{subsubsec:SBR}
These methods operate on either the original graph or extracted subgraphs, leveraging graph algorithms to identify reasoning paths.
Starting from the query-relevant entities, these algorithms iteratively traverse node connections to generate candidate paths, utilizing techniques such as Breadth-First Search (BFS), Depth-First Search (DFS), Dijkstra's algorithm, and their variants. 
The procedure is formally defined as:
\begin{equation*}
\small
\ShortRetrieval(G{~or~}g_q, q, \epsilon_q) \rightarrow \mathcal{P}_{q} = \bigcup\nolimits_{v_i^{(q)} \in \epsilon_q} \mathcal{P}_{(q,v_i^{(q)})}
\end{equation*}
Here, $\mathcal{P}_{q}$ represents the union of the reasoning path sets $\mathcal{P}_{(q,v_i^{(q)})}$, each focusing on an entity \( v_i^{(q)} \) in \( \epsilon_q \). 
Two primary approaches are applied to extract paths for each query entity: \textit{\getAllPaths} and \textit{\getShortestPaths}.

{\it \underline{a) {\GetAllPaths} (EPR).}} EPR {\cite{ref:kelp}} enumerates all possible paths from an entity \( v_i^{(q)} \in \epsilon_q \) to every reachable node within the graph or subgraph. It captures diverse reasoning chains, providing additional context for LLMs, although it may introduce information that is not related to the query, potentially adding noise to the reasoning process of the LLMs.

{\it \underline{b) {\GetShortestPaths} (SPR).}} SPR {\cite{IndexSPR,PullNet,EmbedKGQA,ScaleSPR,DecAF}} identifies all shortest paths from an entity \( v_i^{(q)} \in \epsilon_q \) to reachable nodes, ensuring that the extracted paths are concise and directly relevant
to the query.

\subsubsection{{\bf One-way Semantic-Augmented Retrieval ({\bf OSAR})}}
\label{subsubsec:OSAR}
Similar to the SAE methods in the {\preRetrieval} module, the OSAR methods leverage semantic models to evaluate and select the \( N_p \) most relevant paths from a candidate path set \( \mathcal{P}\in G \text{~or~} g_q \),\footnote{$\mathcal{P}$ is typically obtained from SBR methods, such as EPR or SPR.} to construct a refined set of reasoning paths \( \mathcal{P}_{q} \) that are most relevant to the query \( q \).
Path semantic relevance can be evaluated using either EEMs or LLMs, as outlined in Section~\ref{subsec:sae}.
Note that when using LLMs to evaluate and select reasoning paths, it may not be feasible to input all paths in \(\mathcal{P}\) in a single call.
In such cases, random selection or EEMs can be used as a preliminary filter to reduce the number of candidate paths.\footnote{Alternatively, multiple LLM calls can be used to process all reasoning paths, improving quality at the cost of efficiency. 
This feature is supported in our implementation (see technical report B.6 and our open-source repository).
}

\subsubsection{{\bf Interactive Semantic-Augmented Retrieval ({\bf ISAR})}} 
\label{subsubsec:ISAR}

Unlike OSAR, which separates path generation and evaluation, ISAR integrates both within an interactive framework. 
Using interactive search algorithms {\cite{BeamSearch,DiverseBeamSearch,ContrastiveSearch}} combined with semantic models, ISAR directs the search process toward semantically relevant graph regions from the outset. By interactively and dynamically evaluating and prioritizing path extensions based on their relevance to the query, ISAR effectively narrows the search space during execution.

For example, in Beam Search, \footnote{See the technical report B.8 for detailed algorithm} which is widely used in GraphRAG workflow~\cite{UHop,ToG}, a fixed number of candidate paths (beams) are explored iteratively to identify the most relevant reasoning path for a query entity $v_i^{(q)} \in \epsilon_q$.
At each step, potential path extensions are evaluated for semantic relevance using a greedy strategy, and the top-$B$ paths are retained. This process continues until a stopping criterion is met, e.g., reaching the maximum path length $L$ or exhausting relevant extensions. The final output consists of the most relevant reasoning paths discovered during the search.

In ISAR, semantic models for evaluating path extensions can be EEMs, LLMs, or a hybrid of both to combine their strengths. Based on practical exploration, we propose three hybrid strategies:
\begin{itemize}[leftmargin=10pt]
\item  {\bf LLMs Refinement (LLMs-R, in ISAR-EEMs):} After the ISAR-EEMs search process, LLMs are used to refine the retrieved paths by reducing redundancy and improving semantic coherence.
     
\item {\bf EEMs Pre-filter (EEMs-PF, in ISAR-LLMs):} At each step of the LLMs-based search, EEMs pre-rank and filter candidate extensions before passing them to the LLMs, reducing input size and improving efficiency.
    
\item {\bf EEMs Supplement (EEMs-S, in ISAR-LLMs):} At each step of path expansion, if the LLMs generate too few or low-quality candidates, EEMs are used to supplement the expansion with top-ranked relevant paths. 
\end{itemize}



Additionally, for small-scale graphs, LLM-agent-based ISAR methods {\cite{DoG}} can generate reasoning paths through single- or multi-turn interactions, if LLMs are fine-tuned on domain-specific knowledge.
Despite domain-specific effectiveness, this approach suffers from limited generalizability and efficiency due to its reliance on fine-tuned LLMs, context length constraints, and poor scalability. Hence, it is not our focus of the framework.

%% file: tables/solutions.tex
\begin{table*}[h]
	\caption{Design Solutions of LEGO-GraphRAG Framework: \it{\small The framework comprises of four groups of instances: (I) SBE \& SBR; (II) SBE \& I/OSAR; (III) SAE \& SBR; (IV) SAE \& I/OSAR. 
 When the subgraph-extraction module is optional, an additional group, (V) SBR or I/OSAR, is included.}}
    \vspace{-65pt}
	\label{tab:design-space}
	\resizebox{\textwidth}{!}{%
		\begin{tikzpicture}
			\tikzstyle{block} = [draw, rounded corners=2mm, thin, minimum width=8cm, minimum height=2.5cm, align=center, inner sep=0pt, text width=8cm, fill=gray!5]

              \tikzstyle{block2} = [draw, rounded corners=2mm, thin, minimum width=8cm, minimum height=1.5cm, align=center, inner sep=0pt, text width=8cm, fill=gray!5]
			\tikzstyle{title} = [align=center, font=\bfseries, text width=8cm]
			\tikzstyle{vtitle} = [align=center, font=\bfseries, rotate=90, anchor=center, text width=6cm]
			\node[title] (title1) at (-8.2, 5) { \textsc{Module I: Subgraph-Extraction}};
			\node[title] (title2) at (0, 5) { \textsc{Module II: Path-Retrieval}};
			\node[vtitle] (title3) at (-13, 4.2) { \textsc{\footnotesize Solution I: \\Structure-based \\Methods}};
			\node[vtitle] (title4) at (-13, 1.6) { \textsc{\footnotesize Solution II: \\Semantic-Augmented \\Methods}};
			
			\node[block2] (block1) at (-8.15, 4) {
				{\bf $\circ$ Structure-Based Extraction (SBE)} \\ 
                {\small
                    -\text{Random Walk-based: Random Walk (RW)}\\
                    -\text{Neighborhood-based: K-hop Subgraph-Extraction (KSE)} \\
				-\text{Structural Importance-based: Personalized PageRank (PPR)} }\\
			};
			
			\node[block2] (block2) at (0.15, 4) {
				{\bf $\circ$  Structure-Based  Retrieval (SBR)} \\ 
                {\small
				-\text{Enumerated Path-Retrieval (EPR)} \\
                    -\text{Shortest Path-Retrieval (SPR)} 
                    }
			};
			
			\node[block] (block3) at (-8.15, 1.8) {
				{\bf $\circ$ Semantic-Augmented Extraction  (SAE)} \\ 
                {\small
				-\text{with \underline{End-to-End Models (EEMs)}}\\
                -\text{with \underline{Large Language Models (LLMs)} } \\ 
                }
			};
			
			\node[block] (block4) at (0.15, 1.8) {
				{\bf $\circ$ One-way Semantic-Augmented Retrieval (OSAR)} \\ 
                {\small
			-\text{with \underline{End-to-End Models (EEMs)}} \\
            -\text{with \underline{Large Language Models (LLMs)} }\\
            }
				{\bf $\circ$ Interactive Semantic-Augmented Retrieval (ISAR)} \\ 
		      {\small
            -\text{with \underline{End-to-End Models (EEMs)}}\\
            -\text{with \underline{Large Language Models (LLMs)} }
            }
            \\ };
			
			\draw[rounded corners=2mm, thick] 
			(-12.4, 0.4) rectangle (4.4, 5.2);
			
			\draw[thick] (-12.4, 3.15) -- (4.4, 3.15);
		
			\draw[thick] (-4, 0.39) -- (-4, 5.2);
		\end{tikzpicture}
        	}
            
            \vspace{-60pt}           
\end{table*}

%% file: tables/pipelines.tex




\begin{table*}[]
    \footnotesize
        \caption{Five Groups of Instances under the LEGO-GraphRAG Framework}
        \vspace{-10pt}
        \label{tab:flat_table}
        \resizebox{\textwidth}{!}{%
        \begin{tabular}{
            >{\centering\arraybackslash}m{2.5cm} 
            >{\centering\arraybackslash}m{2.5cm} 
            >{\centering\arraybackslash}m{3cm}   
            >{\centering\arraybackslash}m{3cm}   
            >{\centering\arraybackslash}m{4cm}   
        }
        \hline
        \multicolumn{2}{c}{\textbf{Group}} &
          \textbf{\makecell{Subgraph-Extraction}} &
          \textbf{\makecell{Path-Retrieval}} &
          \textbf{\makecell{Implemented Instances}} \\ 
        \hline
        \multicolumn{2}{c}{\textbf{\makecell{Structure-based Methods on Both \\Modules (Group (I): {\it SBE \& SBR}) }}} &
          SBE (-RW/KSE/PPR) &
          SBR (-EPR/SPR) &
          Our Instance: No.{\PPRxSPR} \\ 
        \hline
        \multicolumn{2}{c}{\multirow{2}{*}{\textbf{\makecell{Semantic-Augmented Methods\\ on Both Modules \\ (Group (II): {\it SAE \& I/OSAR})}}}} &
          SAE (-EEMs/LLMs) &
          OSAR (-EEMs/LLMs) &
          \makecell{Our Instances: No.{\EEMxEEMxOSAR}, {\EEMxLLMxOSAR}, {\LLMxEEMxOSAR}, {\LLMxLLMxOSAR}} \\ \cline{3-5} 
        \multicolumn{2}{c}{} &
          SAE (-EEMs/LLMs) &
          ISAR (-EEMs/LLMs) &
          \makecell{GCR (arXiv24)~\cite{GCR} \\ Our Instances: No.{\EEMxEEMxISAR}, {\EEMxLLMxISAR}, {\LLMxEEMxISAR}, {\LLMxLLMxISAR}} \\ 
        \hline
        \multicolumn{2}{c}{\textbf{\makecell{Semantic-Augmented Methods\\ on Subgraph-Extraction \\(Group (III): {\it SAE \& SBR})}}} &
          SAE (-EEMs/LLMs) &
          SBR (-EPR/SPR) &
          \makecell{RoG (ICLR24)~\cite{RoG} \\ GSR (EMNLP24)~\cite{LessIsMore} \\ Our Instances: No.{\EEMxSPR}, {\LLMxSPR}} \\ 
        \hline
        \multicolumn{2}{c}{\multirow{2}{*}{\textbf{\makecell{Semantic-Augmented Methods\\ on Path-Retrieval \\ (Group (IV): {\it SBE \& I/OSAR})}}}} &
          SBE (-RW/KSE/PPR) &
          OSAR (-EEMs/LLMs) &
          \makecell{Our Instances: No.{\PPRxEEMxOSAR}, {\PPRxLLMxOSAR}} \\ \cline{3-5} 
        \multicolumn{2}{c}{} &
          SBE (-RW/KSE/PPR) &
          ISAR (-EEMs/LLMs) &
          \makecell{StructGPT (EMNLP23)~\cite{StructGPT} \\ Our Instances: No.{\PPRxEEMxISAR}, {\PPRxLLMxISAR}} \\ 
        \hline
            \multicolumn{2}{c}{\multirow{3}{*}{\textbf{\makecell{Without Subgraph-Extraction \\ Modules \\(Group (V): {\it SBR or I/OSAR})} }}} &
          None &
          SBR (-EPR/SPR) & -
           \\ \cline{3-5} 
        \multicolumn{2}{c}{} &
          None &
          OSAR (-EEMs/LLMs) &
          \makecell{KELP (ACL24)~\cite{ref:kelp}} \\ \cline{3-5} 
        \multicolumn{2}{c}{} &
          None &
          ISAR (-EEMs/LLMs) &
          \makecell{ToG (ICLR24)~\cite{ToG}, DoG (arXiv24)~\cite{DoG}} \\ 
        \hline
        \end{tabular}%
        \vspace{-10pt}
        }
    \end{table*}

%% file: section/experiment.tex
\section{Experiment}
\label{section: experiment}

Building upon the LEGO-GraphRAG framework, we develop an implementation that facilitates the seamless integration of modules and methods for constructing GraphRAG instances.\footnote{Our implementation is built in Python and integrates libraries such as Transformers {\cite{transformers}}, iGraph {\cite{iGraph}}, PyTorch{\cite{PyTorch}}, and vLLM {\cite{vLLM}} to support flexible configuration of algorithms, models, and reasoning. The semantic models used are publicly available on the Hugging Face {\cite{huggingfaceHuggingFace}}.}
To assess the effectiveness of the LEGO-GraphRAG framework and our implementation, we create versions of our framework based on the core ideas of three state-of-the-art instances—RoG~\cite{RoG}, KELP~\cite{ref:kelp}, and ToG~\cite{ToG}—and align the experimental setups to ensure a fair comparison. As shown in Table~\ref{table:CompareSOTA}, the instances implemented within our framework exhibit performance comparable to their original counterparts across two datasets. 
For KELP, our implementation outperforms the original due to the effective integration of two modules and the optimization of the employed methods.

As shown in Table~\ref{tab:flat_table}, the existing instances are far from covering the four groups of GraphRAG instances within our framework. Thus, we construct {\InstanceNum} distinct GraphRAG instances (numbered and labeled in Table~\ref{tab:flat_table}) by selecting representative methods from {\preRetrieval} and {\retrieval} modules for a detailed empirical study (Section~\ref{exp:instance}). In addition, we conduct separate evaluations of the various methods, strategies, and semantic models utilized in the two modules (Sections~\ref{exp:PreRetrieval} and~\ref{exp:Retrieval}).

\input{tables/CompareSOTA.tex}

\subsection{Experiment Settings}



\textbf{Graph and GraphRAG Query Datasets.} 
We leverage Freebase~\cite{Freebase}, a large-scale, multi-domain knowledge base (e.g., on finance, law, sports) widely used in GraphRAG research~\cite{StructGPT,UniKGQA,KD-CoT,DecAF,RoG,ToG,GNN-RAG,ChatKBQA,GCR,LessIsMore}, along with four established query datasets, WebQSP~\cite{webqsp}, CWQ~\cite{CWQ}, GrailQA~\cite{GrailQA}, and WebQuestions~\cite{UHop}, covering diverse and challenging GraphRAG scenarios.
Following standard settings~\cite{NSM,ToG,UniKGQA,RoG}, we construct dataset-specific graphs, including all triples reachable within the maximum reasoning hops from query entities (100M nodes, 300M edges) and sample 1,000 test queries per dataset (one-hop: multi-hop is 1:1). All experiments (Sections 4.2–4.4) are conducted based on these datasets.
We also include MetaQA~\cite{MetaQA}, a non-Freebase dataset built on Wiki-Movies~\cite{ref:wikimovie}, a single-domain base focused on movies (40K nodes, 130K edges), with 1,000 test queries to assess cross-base and cross-domain transferability  of our framework.\footnote{Further details of datasets  are in our technical report B.2 and B.3.}


\interfootnotelinepenalty=10000

\textbf{Metrics.} Our empirical study is based on three key metrics: quality, efficiency, and cost.
{\it \underline{a) Quality.}} 
\uwave{{\it a).1 For the subgraph-extraction}}, F1 Score serves as the primary evaluation metric, balancing Precision and Recall to reflect overall extraction quality.\footnote{
$
\text{{Precision}} = |\mathbf{C}_q \cap \mathbf{A}_q| / |\mathbf{C}_q|
$,
$
\text{{Recall}} = |\mathbf{C}_q \cap \mathbf{A}_q| / |\mathbf{A}_q|
$,
where \( \mathbf{C}_q \) and \( \mathbf{A}_q \) represent the predicted and ground truth sets, respectively. The F1 score is the harmonic mean of Precision and Recall:
$
\text{F1 score} = (2 \cdot \text{{Precision}} \cdot \text{{Recall}}) / (\text{Precision} + \text{Recall})
$
\label{footnote:f1}
} 
However, since this module directly influences the retrieval space for downstream modules, we introduce a minimum Recall threshold to ensure sufficient coverage of relevant entities and relations. 
Relying solely on high F1 Score may favor Precision at the cost of missing important components.
In practice, a fixed Recall threshold (e.g., around 60\%) is usually sufficient to ensure downstream performance, after which F1 Score can guide subgraph-extraction.
We also explore adaptive thresholding based on query complexity, graph density, downstream tasks, and historical logs, balancing coverage and efficiency.\footnote{
Detailed strategies are available in our technical report B.5. }
\uwave{{\it a).2 For the path-retrieval}}, the F1 score\footref{footnote:f1} measures the alignment between the retrieved reasoning paths and the ground truth answers. For sets containing the same number of reasoning paths, a higher F1 score indicates better quality.
\uwave{{\it a).3 For the GraphRAG instance}}, we adopt Hits@1 as the main evaluation metric, following prior work~\cite{knowledgeaugmented, StructGPT, CoK, KB-BINDER, ToG, ToG2.0}. 
It measures the proportion of outputs that exactly match the ground truth.\footnote{Note the gap between the evaluation of path-retrieval and that of the GraphRAG instance, which arises from the ability of LLMs to  utilize the reasoning paths.} We also report two commonly used complementary metrics—F1 score and LLM-based evaluation. Both show similar trends to Hits@1 across different GraphRAG instances and do not affect our experimental conclusions.\footnote{Detailed results are provided in the technical report B.4.}
\underline{b) Efficiency.} We record the runtime of both modules in the GraphRAG instance per query. 
\underline{c) Cost.} We track token cost and peak GPU memory usage of the GraphRAG instance on EEMs and LLMs, reflecting processing load and computational resource demands, respectively.

\textbf{Settings for Graph Instance Experiments.} For GraphRAG instances in this study, we selected methods and semantic models that balance efficiency and quality within their respective categories, based on experiments on subgraph-extraction and path-retrieval modules (Sections~\ref{exp:PreRetrieval} and~\ref{exp:Retrieval}).
To ensure fair comparisons, we adjusted parameters within each instance so that all instances ultimately retrieved the same number of reasoning paths.\footnote{In this paper, we follow prior work {\cite{RoG,GNN-RAG,LessIsMore}} by setting the number of reasoning paths to 32. Note that for instances integrating LLMs in OSAR and ISAR, the number of reasoning paths may be fewer than 32 due to the uncertainty in their outputs.} The key experimental settings are as follows, and detailed settings for each instance are provided in supplemental material:
{\it \underline{a) EEMs\&LLMs.}} A sentence-Transformer (ST)~\cite{MiniLM} and {\qwen{2}{72}}~\cite{qwen2} are employed as the EEMs and LLMs implementations, respectively. Note that for all experiments in this paper, the batch size of EEMs is set to 64 and the max input token of LLMs is set to 16k.
{\it \underline{b) SBE.}} The PPR algorithm is applied, with a maximum of nodes $N_{ppr}=1,000$ retained.
{\it \underline{c) SAE.}} Edge pruning (EP) is used for EEMs and LLMs based on subgraphs extracted by PPR, with a maximum of edges $N_{e}=64$ retained.
{\it \underline{d) SBR.}} The Dijkstra algorithm is employed for SPR implementation.
{\it \underline{e) OSAR.}} Select $N_p=32$ reasoning paths from those retrieved using SPR, leveraging both EEMs and LLMs.
{\it \underline{f) ISAR.}} The Beam Search algorithm is selected for implementation and combined with both EEMs and LLMs. The beam width is set to $B=8$, and the maximum path length retained is $L=4$.
{\it \underline{g) Generation.}} We utilize various LLMs with differing architectures and scales~.\footnote{Glm4-9B~\cite{glm2024chatglm}, \llama{3.3}{70}~\cite{llama3}, \qwen{2}{7} and \qwen{2}{72}~\cite{qwen2}}


\textbf{Settings for Module Experiments.}
To evaluate the quality and efficiency of methods in subgraph-extraction and path-retrieval modules, we implement representative methods from each method category, and conduct independent experiments. The settings are as follows:
{\it \underline{a) EEMs\&LLMs Selection.}} A variety of EEMs, including statistical model, embedding model, and reranker, are employed. Specifically, we adopt BM25 as a representative statistical model, a sentence-Transformer (ST)~\cite{MiniLM} as a widely used embedding model, BGE-Reranker (BGE)~\cite{BGE} as an effective re-ranker, and \qwen{2}{72}~\cite{qwen2} as the LLM implementation.

{\it \underline{b) Settings for the Subgraph-Extraction Module.}}
\uwave{{\it b).1 SBE.}} We employ PPR as a structural importance-based method, RW as a random walk-based algorithm, and KSE as a neighborhood-based strategy.
\uwave{{\it b).2 SAE.}} Node Pruning (NP), Edge Pruning (EP), and Triple Pruning (TP) are applied across all selected EEMs and LLMs, based on subgraphs extracted by SBE, with PPR selected due to its superior quality and efficiency in our evaluations.

{\it \underline{c) Settings for the Path-Retrieval Module.}}
We extract 3,000 subgraphs from four evaluation datasets using methods implemented in the subgraph-extraction module. From each dataset, 250 test queries (1,000 total) are sampled, and three subgraphs per query are generated using SBE, SAE-EEM, and SAE-LLM, with the most effective implementations selected based on our experiments. These subgraphs are used to evaluate the methods in the path-retrieval module independently. The settings for each method are as:
\uwave{{\it c).1 SBR.}} The Dijkstra algorithm is employed for SPR, while the DFS algorithm is employed for EPR.
\uwave{{\it c).2 OSAR.}} We use all selected EEMs and LLMs to perform semantic evaluation and refinement on the paths obtained through SPR (a superior SBR method based on our evaluation).
\uwave{{\it c).3 ISAR.}} We combine beam search algorithm (\( B = 8 \) and \( L = 4 \)) with all selected EEMs and LLMs.

%% file: tables/CompareSOTA.tex
\begin{table}[t]
\footnotesize
    \centering
    \caption{Existing Instances vs. LEGO-GraphRAG Instances}
    \label{table:CompareSOTA}
    \vspace{-10pt}
    \resizebox{\columnwidth}{!}{%
    \begin{tabular}{lcccc}
    \toprule
    \multicolumn{1}{c}{\multirow{2}{*}{GraphRAG Instances}} & \multicolumn{2}{c}{WebQSP} & \multicolumn{2}{c}{CWQ} \\\cline{2-5} 
    \multicolumn{1}{c}{}                         & Hits@1       & Recall      & Hits@1     & Recall     \\  \midrule
    RoG~\cite{RoG} (RoG planning w/ChatGPT)                                          & 81.51         & 71.60       & 52.68      & 48.51      \\
    LEGO-RoG (RoG planning w/ChatGPT)                                     & 82.79        & 64.41       & 56.06      & 49.76\\ \midrule
    KELP~\cite{ref:kelp} (one-hop w/gpt-4o-mini)                                          & 31.06        & -       & 14.16      & -      \\
    LEGO-KELP (one-hop w/gpt-4o-mini)                                     & 77.36        & 63.99       & 48.65      & 43.88      \\ \midrule
    ToG~\cite{ToG} (w/\llama{3}{8})                                          & 59.76         & 43.05           & 36.97       & 32.69          \\ 
    LEGO-ToG (w/\llama{3}{8})                                     & 66.44        & 44.77           & 40.26      & 33.63          \\ \bottomrule
    \end{tabular}
    
    }
    \vspace{-10pt}
\end{table}

%% file: section/results.tex
\begin{figure*}[t] \centering
	\includegraphics[width=1\textwidth]{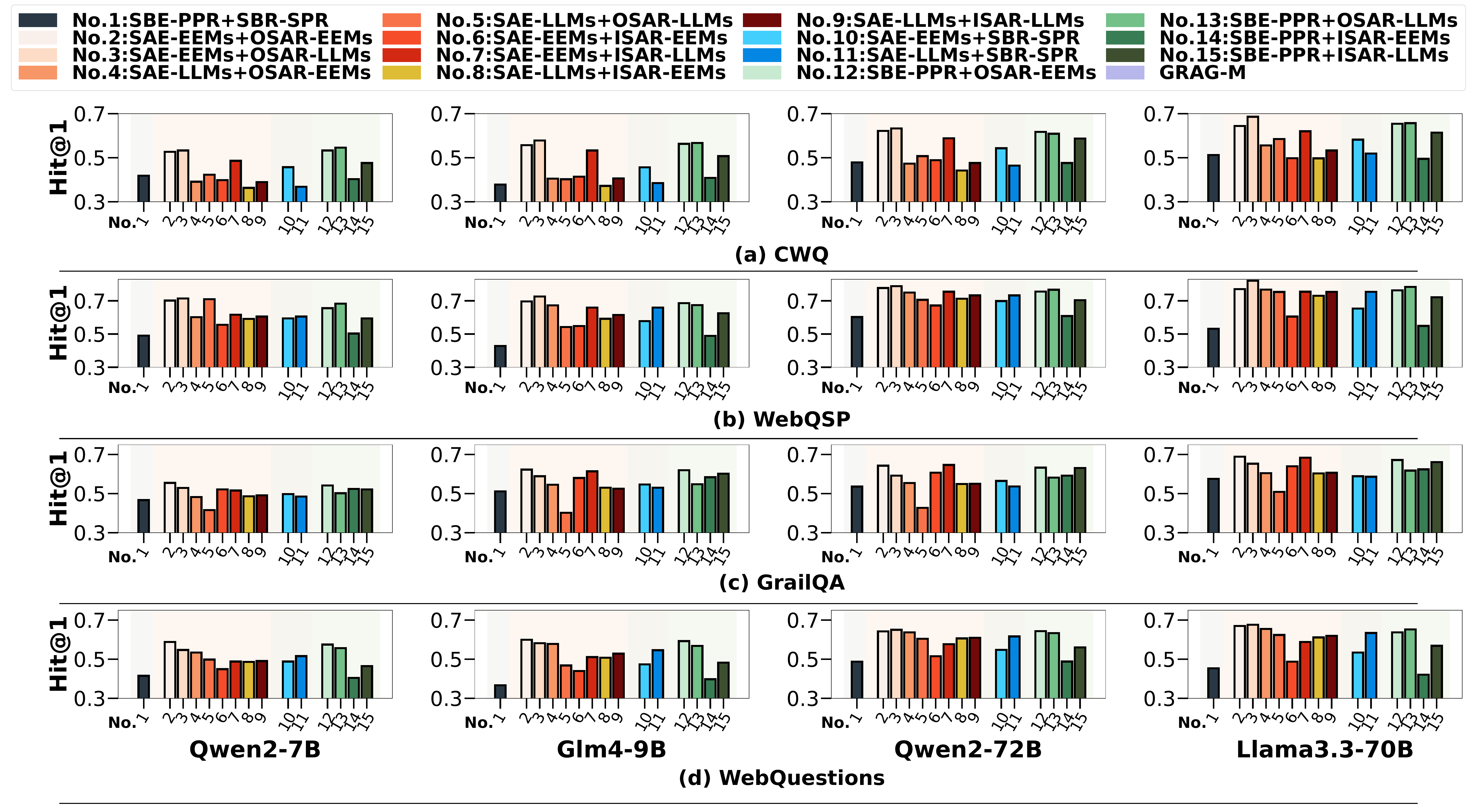}
    \vspace{-20pt}
	\caption{\small Reasoning Results of LEGO-GraphRAG Instances Across Four Datasets} 
	\label{fig:Instance}
\end{figure*}



\begin{figure*}[h!]
    \centering
    \begin{minipage}[b]{0.72\textwidth}
    \centering
    \vspace{-10pt}
    \includegraphics[width=1\textwidth]{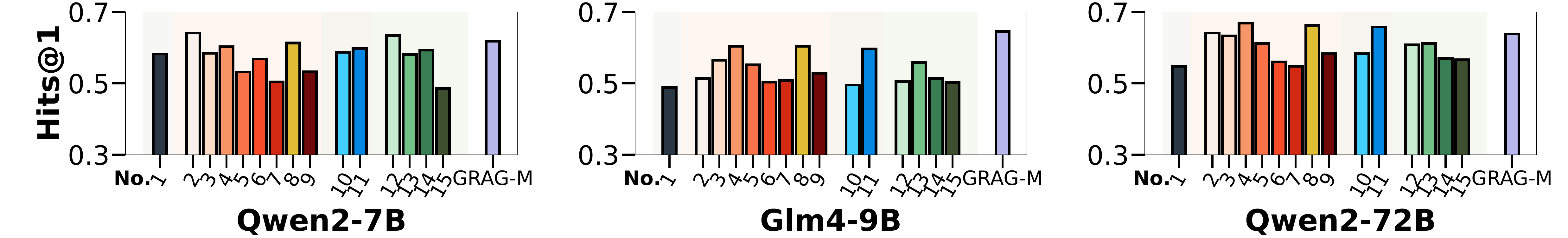}
        \vspace{-20pt}
    \caption{\small Results of LEGO-GraphRAG Instances  and GraphRAG-M Instance on MetaQA Dataset} 
    \label{fig:metaqa}
    \end{minipage}%
    \hspace{0.01\textwidth} 
    \begin{minipage}[b]{0.25\textwidth}
    \centering
    \vspace{-10pt}
    \includegraphics[width=1\textwidth, height = 2.1cm]{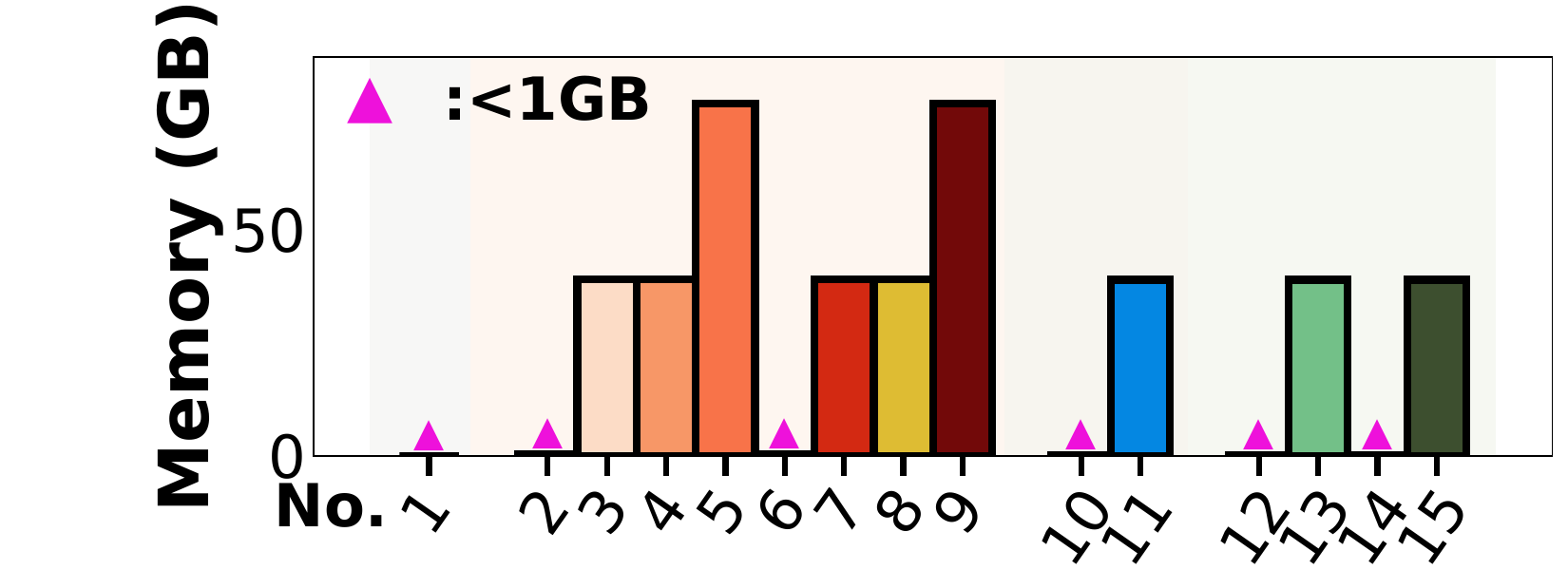}
    \vspace{-20pt}
    \caption{\small Peak GPU Memory Usage} 
    \label{fig:GPU}
    \end{minipage}
    \vspace{-10pt}%
\end{figure*}
\vspace{-10pt}
\begin{figure*}[h!]
    \centering
    \begin{minipage}[b]{0.47\textwidth}
        \centering
        \includegraphics[width=\textwidth]{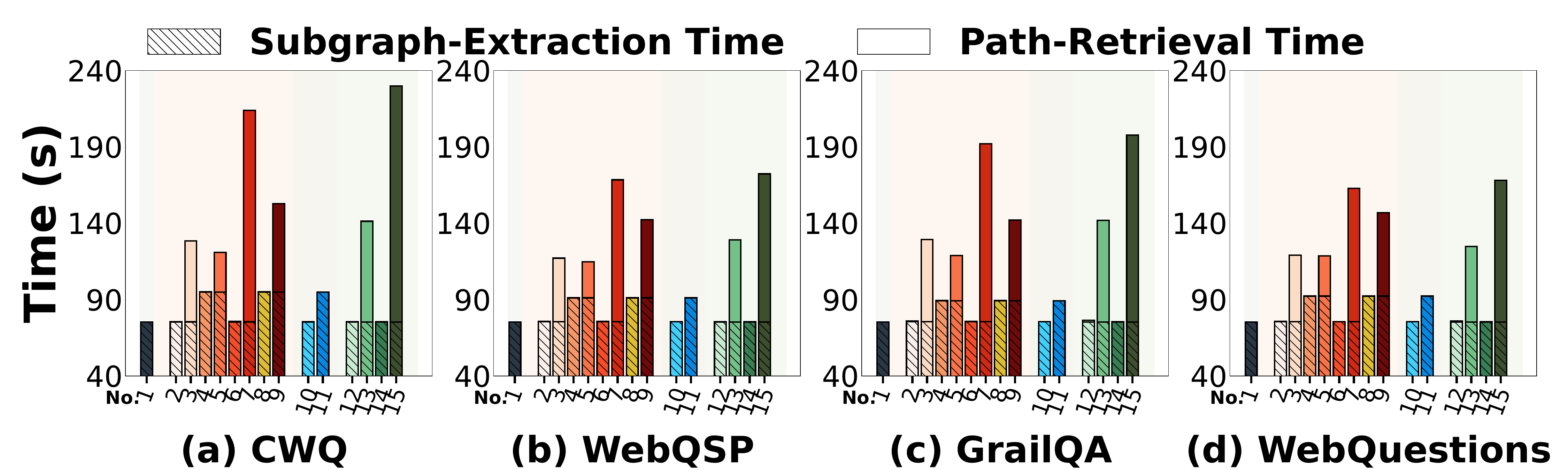}
        \vspace{-20pt}
        \caption{\small Runtime of SE and PR Modules for GraphRAG Instances}
        \label{fig:Instance-time}
    \end{minipage}%
    \hspace{0.01\textwidth} 
    \begin{minipage}[b]{0.47\textwidth}
        \centering
        \includegraphics[width=\textwidth]{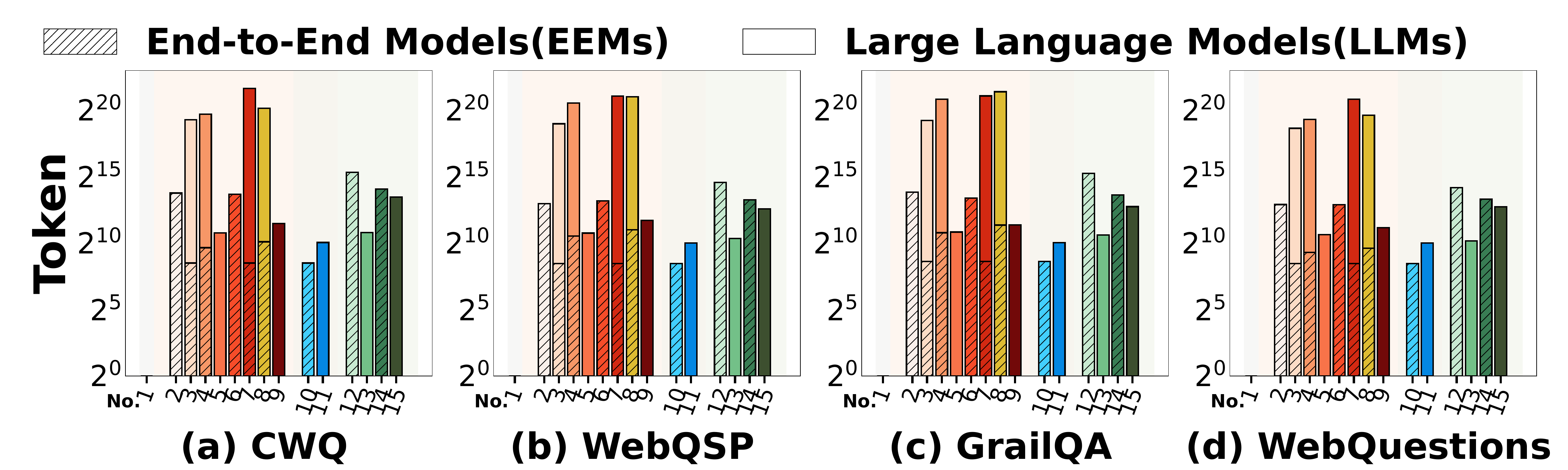}
         \vspace{-20pt}
        \caption{\small Token Costs for EEMs and LLMs in GraphRAG Instances}
        \label{fig:Instance-token}
    \end{minipage}%
    \vspace{-10pt}
\end{figure*}

\begin{figure}[h!]  
    \centering  
    \begin{subfigure}{0.48\textwidth}  
        \centering  
        \includegraphics[width=\textwidth]{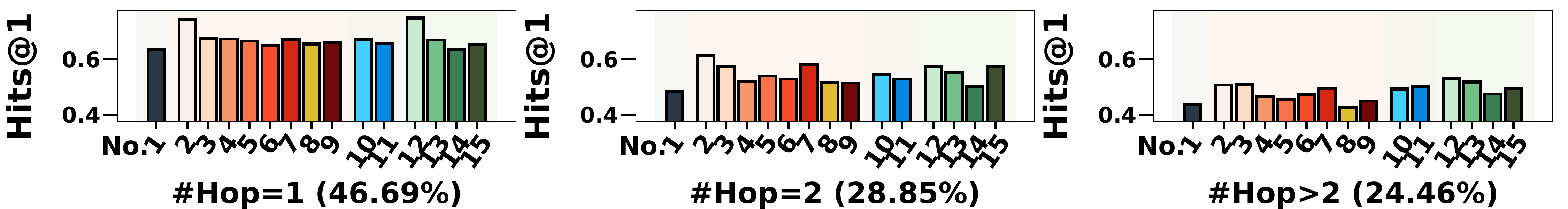}  
        \caption{\small Queries with Varied Number of Reasoning Hops} 
        \label{fig:Hop}  
    \end{subfigure}  
    \begin{subfigure}{0.48\textwidth}  
        \centering  
        \includegraphics[width=\textwidth]{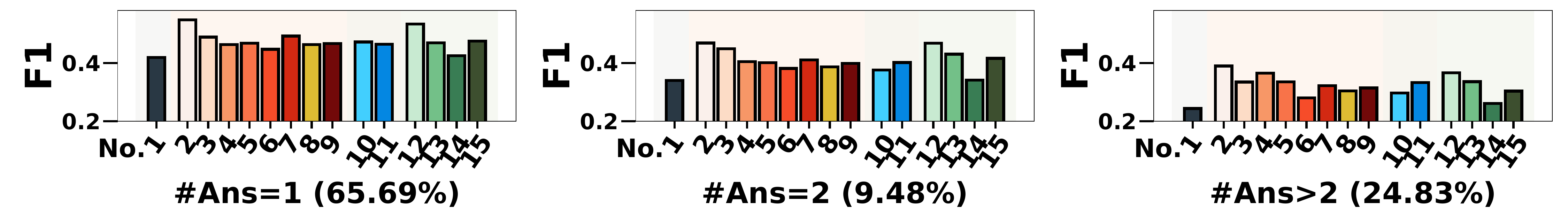}  
                \caption{\small Queries with Varied Number of Answers}
        \label{fig:Answer}  
    \end{subfigure}  
    \begin{subfigure}{0.48\textwidth}  
        \centering  
        \includegraphics[width=\textwidth]{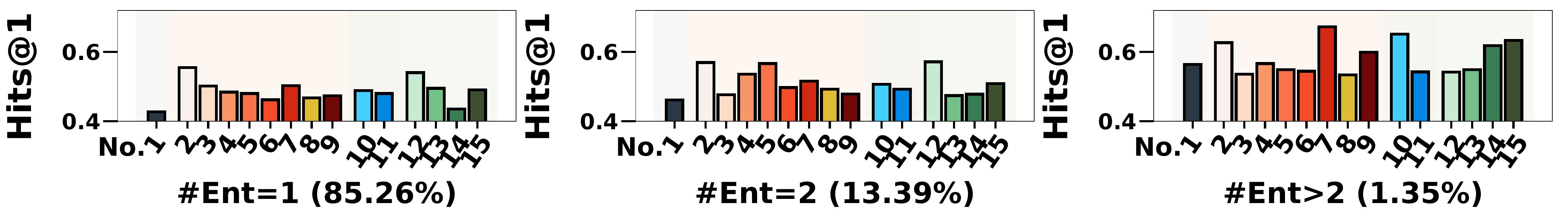}
                \caption{\small Queries with Varied Number of Entities}
        \label{fig:Entity}  
    \end{subfigure}  
    \vspace{-22pt}
    \caption{Instance Performance w.r.t. Queries}
    \label{fig:combined}  
   \vspace{-8pt}
\end{figure} 

\begin{figure*}[t]
    \centering
    \begin{minipage}[b]{0.3\textwidth}
        \centering
        \includegraphics[width=\textwidth]{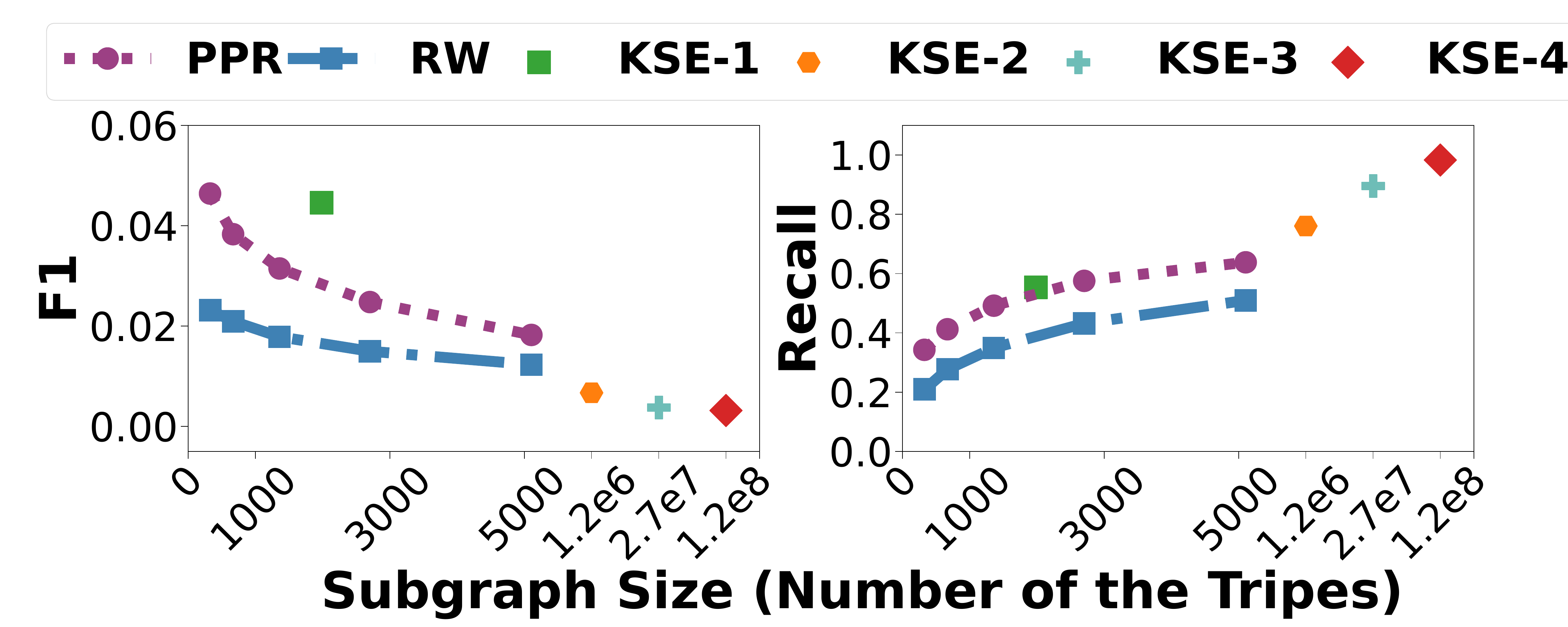}
        \vspace{-20pt}
        \caption{\small Results w.r.t. Subgraph Size (SBE)}
        \label{fig:SE-structmodel}
    \end{minipage}%
    \hspace{0.01\textwidth}
    \begin{minipage}[b]{0.3\textwidth} 
        \centering
        \includegraphics[width=\textwidth]{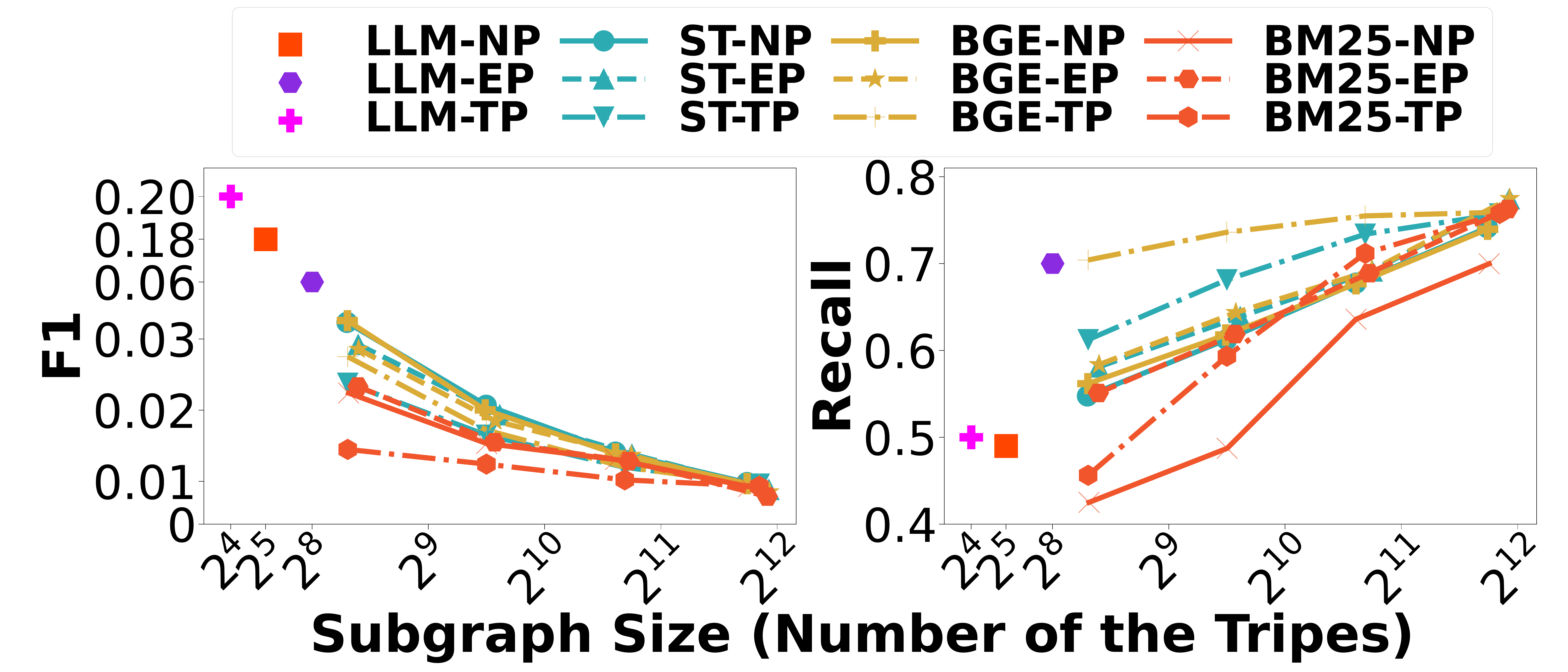}
        \vspace{-20pt}
        \caption{\small Results w.r.t. Subgraph Size (SAE)}
        \label{fig:se-smallmodel}
    \end{minipage}%
    \hspace{0.01\textwidth}
    \begin{minipage}[b]{0.35\textwidth} 
        \centering
        \includegraphics[width=\textwidth]{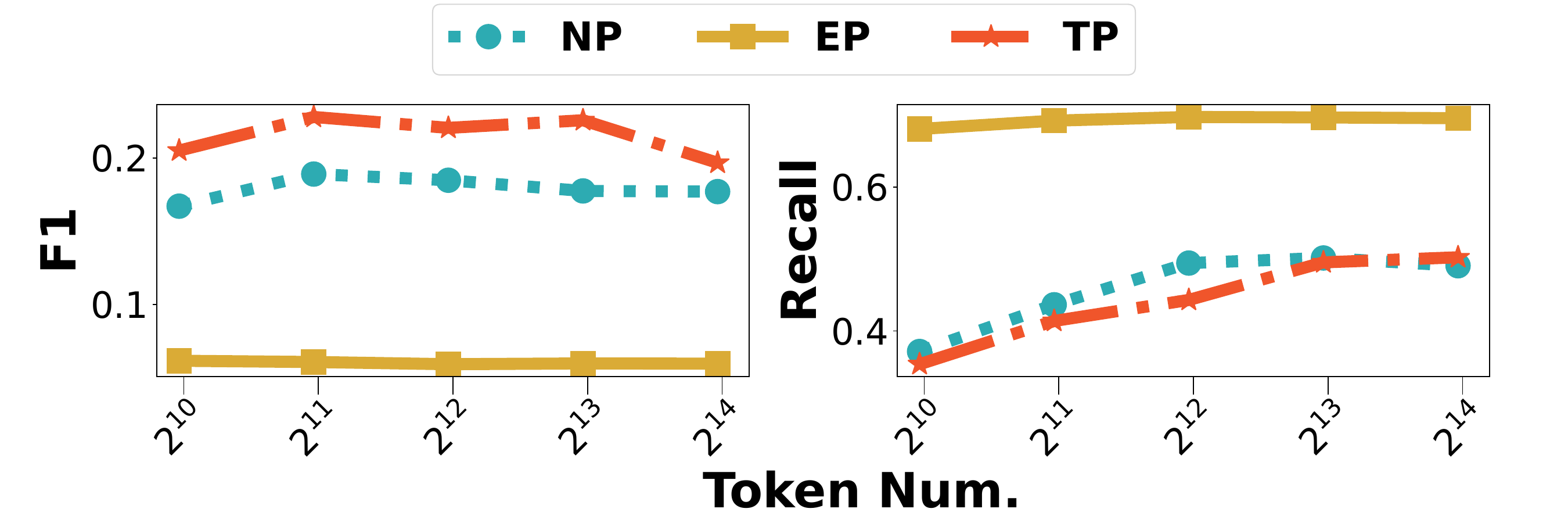}
        \vspace{-20pt}
        \caption{\small Results w.r.t. Token Cost (SAE-LLMs)}
        \label{fig:SE-LLM-Token}
    \end{minipage}%
\end{figure*}

\subsection{Evaluation of GraphRAG Instances}
\label{exp:instance}
We first present the results of all GraphRAG instances across various metrics and analyze key observations.\footnote{We supplement the evaluation with a case study comparing representative reasoning paths and answers (see technical report B.12).}


{\it \underline{a) Reasoning Performance.}} 
As shown in Figure~\ref{fig:Instance}, the reasoning performance of different GraphRAG instances across four datasets follows a consistent trend. With the increase in reasoning model size and capabilities (i.e., from 7B LLMs to 72B LLMs), all instances show significant performance improvements. Next, we analyze the instances according to the groups defined in Table~\ref{tab:flat_table}.

\uwave{{\it a).1 Structural Methods on Both Modules (Group (I)).}} 
Instance \\ No.1 of Group (I) exhibits the worst overall performance.
For example, its Hits@1 is only 0.49 (WebQSP with Qwen2-7B), while  most other instances are close to or exceed 0.6. 
This indicates that structure-based methods alone yield baseline solutions, necessitating semantic models for further improvement.
Notably, Instance No.1 outperforms certain semantically augmented instances on specific datasets, such as No.6, No.8, and No.14 (CWQ with Llama3.3-70B). It shows that while semantic augmented methods generally perform better, exceptions depend on query complexity, domain, and integration of semantic models, as discussed later.

\uwave{{\it a).2 Semantic Augmentation on Both Modules (Group (II)).}} 
In \\ Group (II), instances using EEMs for subgraph-extraction (e.g., No.2, 3, 6, 7) consistently outperform those using LLMs (e.g., No.4, 5, 8, 9) across GrailQA, and all reasoning models. 
This indicates that LLMs are unsuitable for subgraph-extraction when both modules are semantically augmented, as they may overly prune graph information, losing critical intermediate information in the reasoning paths, a limitation seen in instances like GCR~\cite{GCR}, which compromises stability and practicality.

Also, Instances No.2 and 3 in Group (II) achieve the best overall performance, consistently ranking in the top-3 in all reasoning models (WebQSP). 
This highlights the effectiveness of combining EEMs for subgraph-extraction with OSAR for path-retrieval, a strategy yet unexplored by existing methods like GSR~\cite{LessIsMore}, which pairs EEMs with structure-based path-retrieval.
Notably, among Group (II) instances using EEMs for subgraph-extraction (i.e., No.2, 3, 6, 7), Instance No.6 performs the worst. For example, its performance is similar to that of Instance No.1, which uses only structure-based methods (CWQ with all reasoning models). 
This suggests that ISAR pairs better with LLMs than EEMs, as EEMs' weaker semantics can yield noisy scores on long paths and small score gaps, making pruning unreliable. 
One way to improve ISAR-EEMs is to enhance the EEMs themselves, for example, through customization or fine-tuning for specific domains or graphs. In addition, we propose two complementary strategies: {\bf triple-level scoring}, which evaluates only the newly added step at each expansion to reduce error accumulation, and {\bf dynamic pruning}, which adjusts the number of retained paths based on score distribution, helping preserve good candidates when scores are close.\footnote{Detailed discussion and results are available in our technical report B.8.}

\uwave{\textit{a).3 Semantic Augmentation on one Module (Group (III)\&(IV)).}}
\\ We observe that the performance of instances in Group IV (i.e., Instances No.12-15) generally surpasses that of Group III (i.e., Instances No.10, 11).
This suggests that if only one module is semantically augmented, prioritizing the path-retrieval module over the subgraph-extraction module is more effective.  
Using structure-based methods for subgraph extraction remains efficient and sufficient, unlike existing instances such as RoG~\cite{RoG} and GCR~\cite{GCR}, which use LLMs (Llama-2-7b, Llama-3.1-8b) for subgraph-extraction and encounter performance and efficiency bottlenecks.
Additionally, Group (II) instances do not consistently outperform Groups (III) and (IV), implying that semantically augmenting both modules is unnecessary in resource-limited contexts.
Moreover, as shown in Figure~\ref{fig:metaqa}, results on MetaQA align with findings from Freebase-based datasets. The small scale and single-domain nature of the graph lead to uniformly high scores across all instances. Notably, Instance No.14 (ISAR-EEMs) performs comparably to or slightly better than Instance No.15 (ISAR-LLMs), suggesting that EEMs can be a viable and more efficient alternative to LLMs in such settings.


{\it \underline{b) Runtime.}} Figure~\ref{fig:Instance-time} shows the average runtime of different GraphRAG instances for subgraph-extraction and path-retrieval modules on a single query.
A key observation is that subgraph-extraction is the primary bottleneck in GraphRAG runtime of all instances. 
For example, Instance No.1, using structure-based methods, requires approximately 70 seconds for subgraph-extraction (i.e., PPR algorithm) but less than 0.1 seconds for path-retrieval.
Since subgraph-extraction is a crucial but time-intensive step in the GraphRAG workflow, it poses significant challenges for the practical development of GraphRAG systems on large-scale graphs like Freebase.
Unfortunately, most existing instances, such as KELP~\cite{ref:kelp}, ToG~\cite{ToG}, and DoG~\cite{DoG}, overlook the efficiency of this process. 

Additionally, in the path-retrieval, methods using LLMs (especially under ISAR) exhibit  longer runtimes compared to those using EEMs.
For example, Instance No.12 (OSAR-EEMs) and No.14 (ISAR-EEMs) complete path-retrieval in under 1.15 seconds and 0.19 seconds, respectively, while Instance No.13 (OSAR-LLMs) and No.15 (ISAR-LLMs) take 66.64 seconds and 122.43 seconds, respectively, on GrailQA. 
This shows that using the ISAR-LLMs methods in the path-retrieval  may result in 
unacceptably low query efficiency.



{\it \underline{c) Cost.}} Figures~\ref{fig:Instance-token} and~\ref{fig:GPU} show the average token overhead (EEMs and LLMs) and peak GPU memory usage (average of four datasets) across GraphRAG instances. 
Instances mixing EEMs and LLMs (e.g., No.3, 4, 7, 8) incur higher token costs than those using the same model type (e.g., No.2, 5, 6, 9), with EEM-only setups being the most economical.
GPU usage remains low (<1GB) for EEM-only instances but rises sharply for LLM-based ones, peaking at ~80GB when LLMs are used in both modules (e.g., No.5, 9). The results underscore the performance–cost trade-offs of semantic augmentation and the importance of corresponding optimization.



{\it \underline{d) Analysis by Query Type.}}  
We analyze performance across different query types 
by varying reasoning hops, number of answers, and number of query entities. As shown in Figure~\ref{fig:Hop}, performance consistently declines with more reasoning hops, especially for structure-based methods (e.g., Instance No.1), which struggle on multi-hop queries. 
Figure~\ref{fig:Answer} shows a similar drop from single- to multi-answer queries, where semantic methods are more resilient.
Figure~\ref{fig:Entity} shows improved performance with more query entities, likely due to increased chances of retrieving relevant paths—even benefiting structure-only instances like No.1.

{\it \underline{e) Integrating Microsoft GraphRAG.}} To evaluate compatibility with Microsoft GraphRAG, we implement an instance (GRAG-M), which follows its workflows: performing community detection on the graph and precomputing textual summaries for each community. 
On the MetaQA dataset, GRAG-M uses the reasoning paths from Instance 12 (which showed strong performance in our experiments) and provides both the paths and summaries as input to the LLMs.
As shown in Figure~\ref{fig:metaqa}, GRAG-M performs well on Qwen2-7B and Glm4-9B. Gains diminish with Qwen2-72B, where the larger model already captures sufficient context, reducing the benefit of precomputed summaries. 
\subsection{Evaluation of {\PreRetrieval} Module}
\label{exp:PreRetrieval}

\begin{figure}[t]
    \centering
    \begin{minipage}[b]{0.24\textwidth}
        \centering
    
        \captionof{table}{\small PPR Runtime  w.r.t. Graph Size}  
        \vspace{-5pt}
        \label{tab:pprtime}
        \resizebox{\linewidth}{!}{%
            \begin{tabular}{l|l|l}
            \toprule
            \textbf{Nodes} & \textbf{Ave. Edges} & \textbf{Ave. Time (s)} \\
            \midrule
            100000    & 431414.5    & 0.58 \\
            1000000   & 5567538.1   & 1.37 \\
            10000000  & 48865879.2  & 11.5 \\
            100176641 & 298458255   & 75.29 \\
            \bottomrule
            \end{tabular}
        }
        \vspace{-5pt}
    \end{minipage}%
    \hspace{0.01\textwidth}
    \begin{minipage}[b]{0.21\textwidth} 
        \centering
        \includegraphics[width=\textwidth]{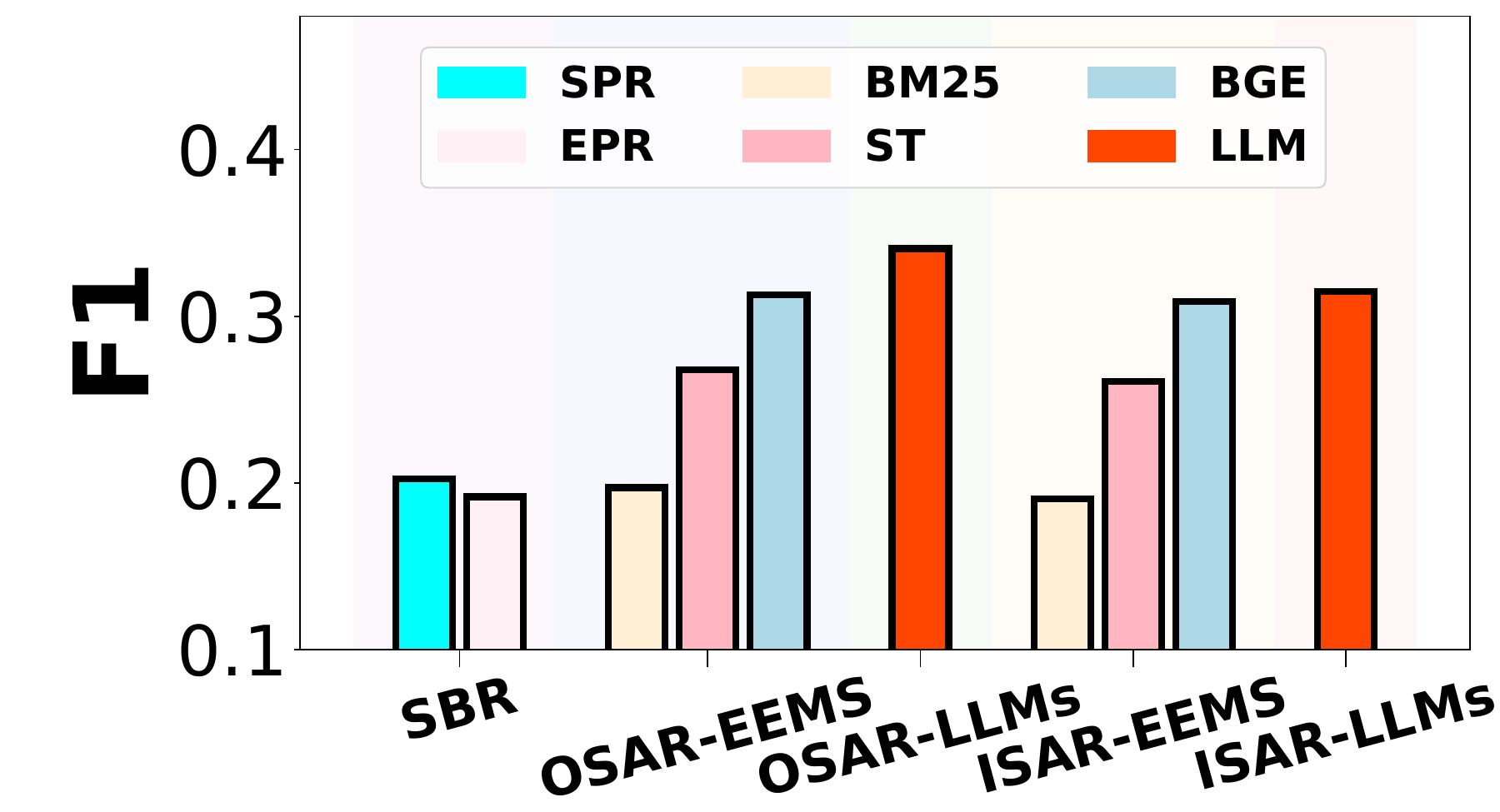}
        \vspace{-21pt}
        \caption{\small F1 Score of Methods in Path-Retrieval Module}
        \vspace{-5pt}
        \label{fig:PR-LLM}
    \end{minipage}%
\end{figure}

{\bf Structure-Based Extraction (SBE).}  
Figure~\ref{fig:SE-structmodel}
shows the variation in F1 score and Recall for the three SBE methods (RW, PPR, and KSE) as the extracted subgraph size changes. The F1 score for all methods decreases as the subgraph size increases, while their Recall consistently improves. Across all subgraph sizes, PPR consistently outperforms RW in both F1 score and Recall, demonstrating the superior effectiveness of PPR over RW. Additionally, RW struggles to achieve a Recall above 60\%, rendering it nearly unusable in GraphRAG.  
Moreover, PPR offers greater flexibility compared to KSE, enabling precise control over subgraph size through hyperparameters. In contrast, KSE relies on hierarchical extraction, which lacks such flexibility.\footnote{Many prior studies {\cite{RoG,GCR,LessIsMore,StructGPT}} first apply KSE to extract 2-hop or 4-hop subgraphs, followed by PPR to distill them into smaller, query-relevant subgraphs.}
These advantages establish PPR as the current optimal solution for SBE. 
As Table~\ref{tab:pprtime} demonstrates, PPR's computational costs become prohibitive for large graphs. 

{\bf Semantic-Augmented Extraction (SAE).}
Figure~\ref{fig:se-smallmodel} shows the Recall and F1 score of different semantic model combinations with three subgraph pruning strategies  under the SAE methods, as the subgraph size varies. 
Among the three semantic models, the BGE and ST significantly surpass BM25 across all pruning strategies and subgraph scales. BGE is slightly better in quality, while ST is even better in efficiency,  as discussed later. 
Considering the trade-off between quality and efficiency, we propose to use ST as the representative EEMs.
For the three pruning strategies, TP exhibits the best quality at smaller subgraph scales, while NP performs the worst. However, as subgraph size increases, the F1 score and Recall of EP, TP, and NP converge. Ultimately, EP slightly outperforms TP, while TP marginally surpasses NP.
Thus, TP is recommended when the size of the extracted subgraph is small; otherwise, EP.


Due to the inherent uncertainty in LLM outputs, SAE-LLMs methods often struggle to control the size of subgraphs extracted, typically producing smaller subgraphs that can reduce path-retrieval quality (see \textbf{Finding 2}).  
Figure~\ref{fig:se-smallmodel} shows the F1 score and Recall for subgraphs generated using SAE-LLMs under three pruning strategies, plotted against the average subgraph size. 
Among the three pruning strategies, EP outperforms NP and TP for SAE-LLMs.
Figure~\ref{fig:SE-LLM-Token} further shows the trends in F1 score and Recall as the number of tokens input into the LLMs varies for NP, TP and EP. Recall for NP consistently surpasses that of both TP and EP, maintaining a gap of around 0.2. Although EP underperforms in terms of F1 score due to the larger subgraphs it extracts, the relatively low Recall of NP and EP may diminish their effectiveness in path-retrieval. As the number of tokens increases, Recall for NP and TP initially rises, then declines, while EP remains nearly constant, indicating that EP achieves more effective subgraph extraction at a lower token cost for the SAE-LLMs methods.
The efficiency of SAE methods is related to the semantic models used. Runtime is influenced by factors such as model parameters, algorithmic complexity, and data volume. In general, the runtime for processing a graph component follows this order: LLMs > BGE > ST > BM25, as shown in Figure~\ref{fig:PR-cost}.

\input{tables/ISAR+EEMs+LLMs}

\subsection{Evaluation of {\Retrieval} Module}
\label{exp:Retrieval}

\begin{figure}[t]
    \centering
    \begin{minipage}{0.23\textwidth}
            \centering
        \includegraphics[width=\textwidth,height=2.3cm]{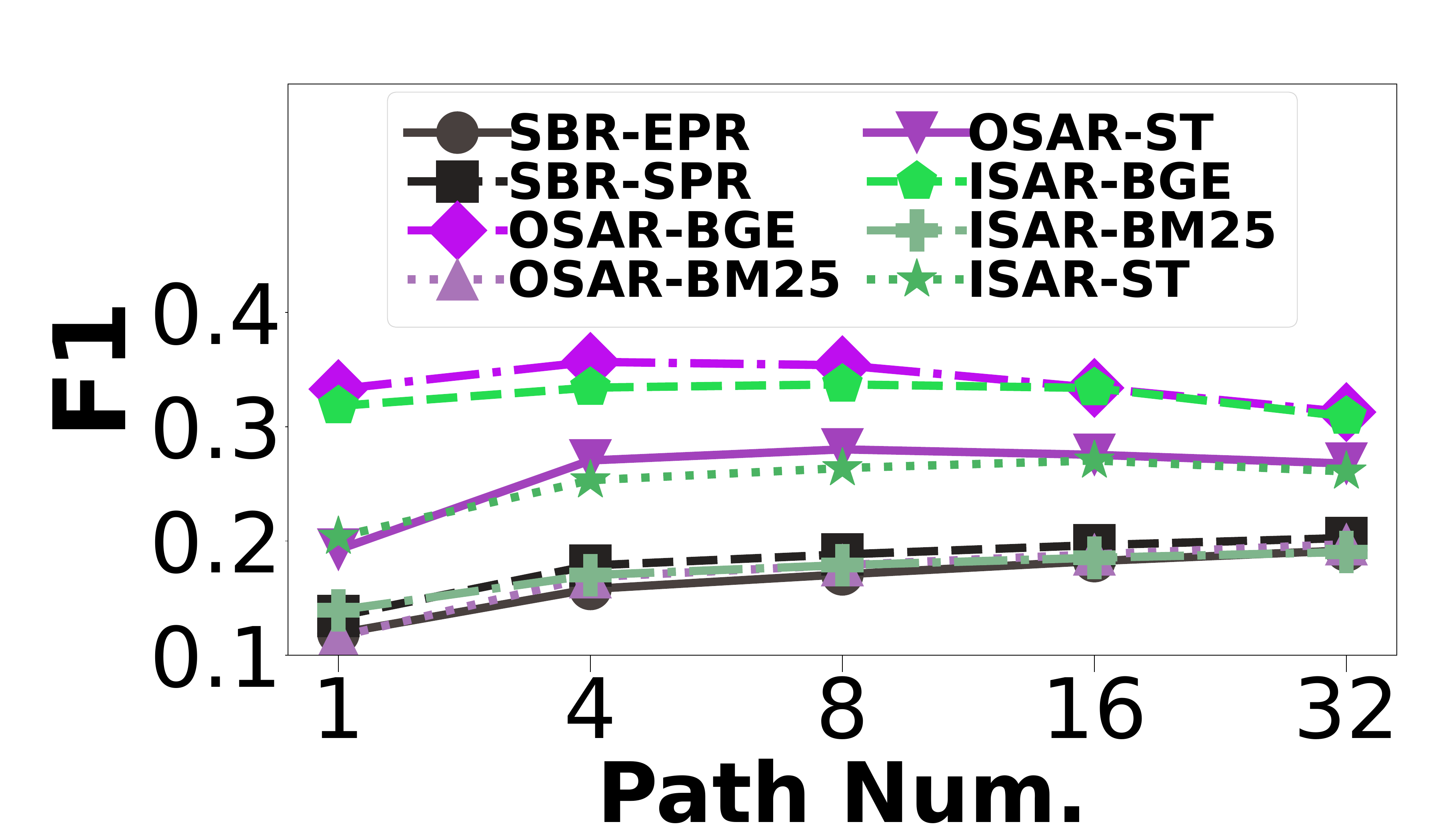}
        \vspace{-20pt}
        \caption{\small F1 Score vs. \# of Paths (for Methods of Path-Retrieval)}
        \label{fig:PR-F1}

    \end{minipage}%
    \hspace{0.1cm}
    \begin{minipage}{0.22\textwidth}
            \centering
        \includegraphics[width=\linewidth,height=2.1cm]{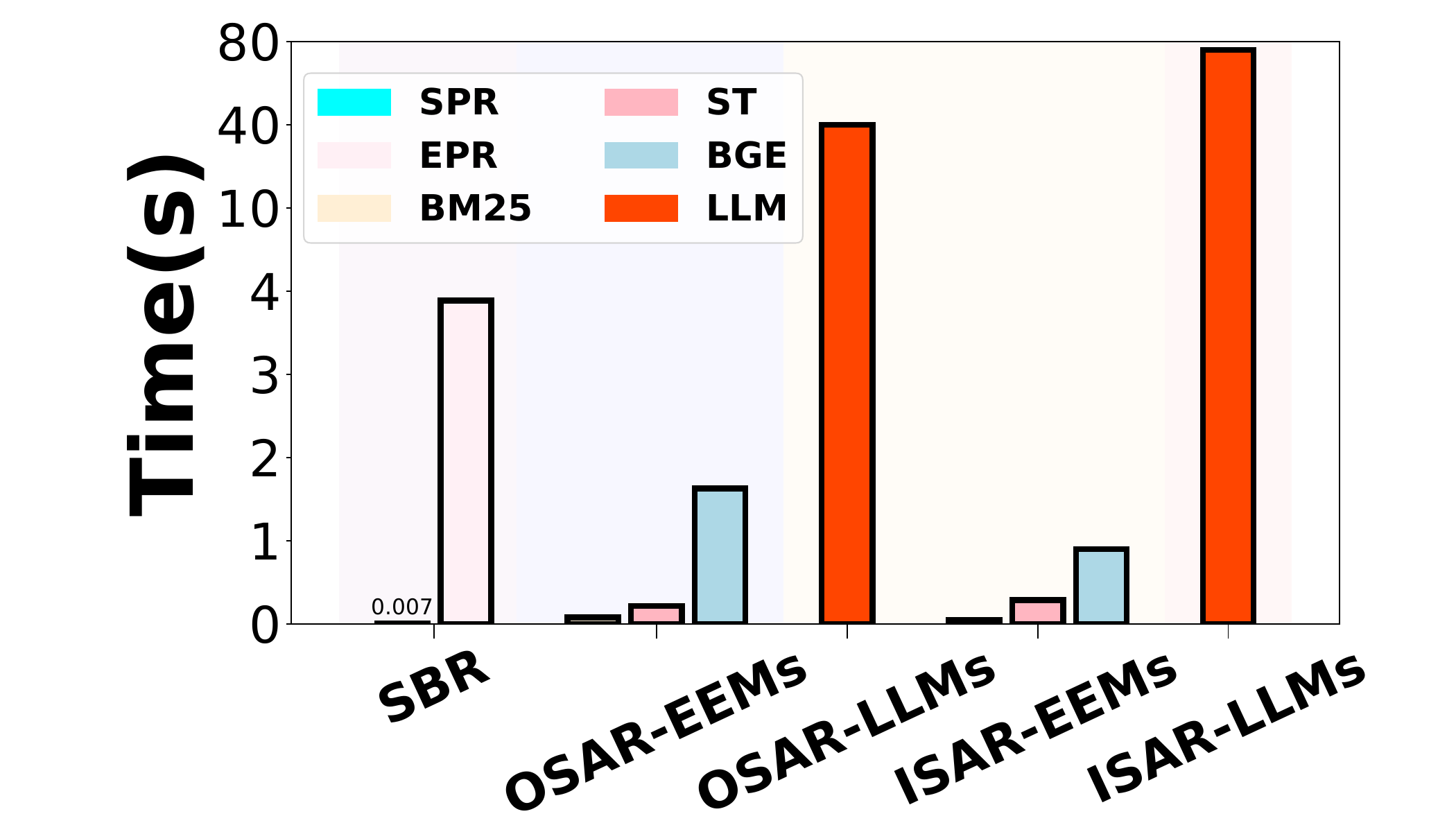}
        \vspace{-14pt}
        \caption{\small Runtime of Methods in Path-Retrieval {\small(WebQSP)}}
        \label{fig:PR-cost}
    \end{minipage}
\end{figure}

Figure~\ref{fig:PR-LLM} shows the F1 score of different methods in path-retrieval module.
   It shows that OSAR outperforms ISAR in all semantic models. SPR method slightly outperforms EPR and I/OSAR with BM25, but underperforms all I/OSAR solutions with ST, BGE or LLM, which suggests that the use of poorer semantic models to aid enhanced retrieval may lead to a loss of quality.
We further investigate the variation in the F1 score of the path-retrieval module when the number of reasoning paths changes, as shown in Figure~\ref{fig:PR-F1}. Since the number of paths output by LLM is inherently uncertain, methods using LLM are excluded.
It shows that the F1 score of SBR and I/OSAR with BM25 gradually increases with the number of paths, yet their F1 score remains low-level, consistently trailing I/OSAR with BGE and ST by approximately 0.1. The gap widens to 0.2 or more when fewer paths are retrieved. In contrast, the performance of I/OSAR with ST and BGE initially improves and then declines as the number of retained paths increases, with OSAR generally outperforming ISAR. Note that F1 scores may be higher for smaller sets of reasoning paths, as they tend to prioritize precision. A slight decrease in F1 for larger reasoning path sets is not indicative of lower quality but rather reflects the broader scope of the retrieved paths.

Figure~\ref{fig:PR-cost} shows the runtime of different methods in the path-retrieval module (WebQSP). Among the SBR solutions, EPR has the longest runtime due to the large number of paths it retrieves, while SPR reduces computation time by retrieving only a single path for each node. 
The EPR-based OSAR method becomes impractical because its excessive data volume results in prohibitive runtime overhead when used with semantic models. As a result, only SPR is employed in this context.
Among the three semantic models, both BGE and ST demonstrate significant quality improvements over BM25 combined with I/OSAR. BGE achieves marginally higher quality, whereas ST exhibits superior efficiency. 

Table~\ref{tab:ISAR_h} reports the results of hybrid ISAR methods combining EEMs and LLMs. All three methods outperform single-model baselines on most datasets, confirming the benefit of combining model strengths. Notably, the EEMs-S strategy yields the best results, suggesting that combining EEMs and LLMs helps reduce the risk of missing important paths during expansion.


%% file: tables/ISAR+EEMs+LLMs.tex
\begin{table}[h]

    \centering  
    \caption{\small Results of Hybrid ISAR Combining EEMs and LLMs}
    \vspace{-10pt}
    \label{tab:ISAR_h}  
    \resizebox{0.46\textwidth}{!}{%
    \begin{tabular}{l|l|l|l|l}  
        \toprule  
        \textbf{Method}
        & \textbf{CWQ}  & \textbf{WebQSP} & \textbf{GrailQA} & \textbf{WebQuestion} \\ \midrule  
                \textbf{SBE+ISAR-EEMs}                         & 0.183         & 0.157           & 0.395             & 0.16                \\

        \textbf{SBE+ISAR-EEMs (LLMs-R)}          & \textbf{0.271}         & \textbf{0.219}           & \textbf{0.379}            & \textbf{0.175}                \\
           \midrule  \midrule  
        \textbf{SBE+ISAR-LLMs}                         & 0.249         & 0.264           & 0.371             & 0.249                \\
        \textbf{SBE+ISAR-LLMs (EEMs-PF)}                           & 0.271         & 0.316           & 0.396            & 0.245                \\
        \textbf{SBE+ISAR-LLMs (EEMs-S)}             & \textbf{0.310} & \textbf{0.373}  & \textbf{0.435}   & \textbf{0.276}       \\      
        \bottomrule 
    \end{tabular}  
    } 
    \vspace{-12pt}
\end{table} 

%% file: section/discussion.tex
\section{Findings and Discussion} 

We summarize key findings from our empirical study and discuss promising ways for addressing GraphRAG's two key bottlenecks: subgraph-extraction efficiency and LLM generation quality.

\subsection{Summary of Key Findings}
{\bf Finding 1: GraphRAG Trade-off Structure.} GraphRAG entails multi-aspect trade-offs among quality (Hits@1), efficiency (runtime), and cost (token and GPU usage). This trade-off spans two modules (subgraph-extraction vs. path-retrieval), two method types (structure-based vs. semantic-augmented), and two types of semantic models (EEMs vs. LLMs).

    \textbf{Finding 2: Module-level Trade-offs.} Subgraph-extraction is the main efficiency bottleneck for large-scale graphs. Overuse of semantic augmentation (e.g., using LLMs) in subgraph-extraction can compromise downstream path-retrieval quality. Targeting semantic augmentation in the path-retrieval module is a more cost-effective strategy to maintain high quality.
        
    \textbf{Finding 3: Method-level Trade-offs.} Structure-based methods offer higher efficiency and lower cost, but semantic-augmented methods are essential for quality, especially for complex (two- or multi-hop) queries. 
    One-way semantic methods (OSAR) balance quality and efficiency better than interactive methods (ISAR).

    \textbf{Finding 4: Model-level Trade-offs.} 
    EEMs are more cost effective in token and GPU usage, while LLMs generally provide better quality, particularly for interactive methods (ISAR). However, LLMs can exhibit unstable generation in some cases, affecting consistency. A balanced approach includes applying LLM-based augmentation in path-retrieval or EEM-based augmentation in both modules.

    \textbf{Finding 5: A Promising yet Underexplored Strategy.} Based on our analysis, the effective approach for GraphRAG would be the one that combines structure based methods for subgraph-extraction with OSAR for path-retrieval. Despite its potential, this strategy remains underexplored in existing works. Also, it opens up opportunities for further optimizations for handling multi-hop and multi-answer queries, as well as improving overall efficiency.

\subsection{Efficiency  of the Subgraph-Extraction}
\label{subsec:eeSE}

We explore promising directions for faster subgraph-extraction, supported by empirical analysis, including limitations, trade-offs, and research opportunities.\footnote{See technical report B.7 for details.} Below, we briefly outline each direction.


{\bf Computational acceleration} yields significant efficiency improvement for SBE methods (e.g., PPR) through 
approximate and distributed algorithms (Table~\ref{tab: PPR-Acceleration}). 
Approximate methods may lower Recall; distributed solutions depend on system setup.

{\bf Precomputation-based acceleration} pre-generates subgraphs (e.g., via clustering or community detection) to scale down the search space for the extraction process in order to reduce runtime overhead. 
As shown in Table~\ref{tab: SE-Acceleration}, our implemented prototype demonstrates the potential of this direction, but it also incurs preprocessing overhead and may limit the ability to access relevant information spread across different subgraphs.

{\bf Vector database-based acceleration} encodes graph components (e.g., nodes or triples) into embedding vectors and stores them in a vector database, thus efficiently identifying query-relevant components based on semantic similarity, followed by subgraph construction centered on them. 
Table~\ref{tab: SE-Acceleration} shows our prototype achieves notable speedups with high Recall. However, the retrieved components often form multiple weakly connected subgraphs, which may hinder multi-hop reasoning.

\input{tables/AccelerationPPR}
\input{tables/PR-LLM}

\subsection{Generation Quality of LLMs}
\label{subsec:LLMsQ}

For the issue of LLMs generating fewer or lower-quality components, we take the path-retrieval module as a example and explore two effective strategies: {\bf Multi-round LLMs (M-LLMs)} 
conduct iterative refinement via repeated LLM calls.
{\bf EEMs supplementation (EEMs-S)} compensates for insufficient outputs by incorporating top-ranked candidate paths from EEMs. 
Table~\ref{tab:PR-LLM} shows both strategies improve performance across most datasets (applied in the SBE+OSAR-LLMs instance), though with added computational overhead of multiple model calls and EEM-based ranking steps.\footnote{More details and analysis are in technical report B.9. }

%% file: tables/AccelerationPPR.tex
\begin{table}[t]
\centering
\caption{{\small Computational acceleration of PPR}} 
\vspace{-10pt}
\resizebox{\linewidth}{!}{%
    \begin{tabular}{l|l|l||l|l}  
        \toprule
        \multicolumn{3}{c||}{\textbf{Approximate PPR (Freebase)}} & 
        \multicolumn{2}{c}{\textbf{Distributed PPR}} \\
        \midrule
        \multicolumn{1}{c|}{\textbf{Method}} & 
        \multicolumn{1}{c|}{\textbf{Recall}} & 
        \multicolumn{1}{c||}{\textbf{Ave. Time (s)}} & 
        \multicolumn{1}{c|}{\textbf{Method}} & 
        \multicolumn{1}{c}{\textbf{Speedup}} \\   
        \midrule
        
        RBS~\cite{RBS} & 0.31 & \textbf{0.51} & 
        HGPA~\cite{HGPA} & 3.4--4.1$\times$ \\   
        
        Fora~\cite{Fora} & 0.18 & 7.61 & 
        PAFO~\cite{PAFO} & 58.7$\times$ \\   
        
        TopPPR~\cite{Topppr} & 0.44 & 42.08 & 
        Delta-Push~\cite{Delta-Push} & \textbf{123--162$\times$} \\   
        
        \midrule
        \textbf{Standard PPR} & \textbf{0.96} & 75.29 & 
        \textbf{Standard PPR} & 1$\times$ (baseline) \\   
        \bottomrule
    \end{tabular}%
}
\label{tab: PPR-Acceleration}
\end{table}

\begin{table}[t]  
\vspace{-10pt}
        \centering   
    \caption{{\small Performance and Cost Analysis of the SE Acceleration Methods on Freebase (About 100M Nodes, 300M Edges)}}  
    \vspace{-10pt}
    \label{tab: SE-Acceleration} 
    \resizebox{0.46\textwidth}{!}{%

    \begin{tabular}{l|l|l|l|l|l}  
        \toprule  
        \textbf{Method}                    & \textbf{Pre-time} & \textbf{Online-time} & \textbf{Recall} & \textbf{F1} & \textbf{WCC} \\   
        \midrule         
        Precomputation                    & 19373s                             & 19.02s                               & 0.52          & 0.0021      & \textbf{1}                                      \\
        Vector Database                    & 14570s                             & \textbf{0.85s}                                & 0.65          & 0.0023      & 12.86                                  \\   
        \midrule  
        \textbf{Standard PPR}                           & \textbf{0s}                                 & 75.29s                      & \textbf{0.96} & \textbf{0.0038} & \textbf{1}                            \\
        \bottomrule  
    \end{tabular}  
    }
\end{table}  

%% file: tables/PR-LLM.tex
\begin{table}[h]  
\vspace{-10pt}
    \centering  
    \small
    \caption{\small Performance of strategies designed to mitigate the limitations of LLM-generated outputs }
    \vspace{-10pt}
    \label{tab:PR-LLM}  
    \resizebox{0.48\textwidth}{!}{%
    \begin{tabular}{l|l|l|l|l}  
        \toprule  
        \textbf{Method} & \textbf{CWQ} & \textbf{WebQSP} & \textbf{GrailQA} & \textbf{WebQuestions} \\  
        \midrule  
        \textbf{SBE+OSAR-LLMs} & 0.304 & 0.401 & \ul{0.401} & 0.328 \\  \midrule 
        \textbf{SBE+OSAR-LLMs (M-LLMs)} & \ul{0.349} & \textbf{0.438} & 0.381 & \ul{0.348} \\  
        \textbf{SBE+OSAR-LLMs (EEMs-S)} & \textbf{0.381} & \ul{0.427} & \textbf{0.476} & \textbf{0.353} \\  
        \bottomrule  
    \end{tabular}  
    }
\end{table}  

%% file: section/conclusion.tex
\section{Conclusion}
\label{section: conclusion}
In this paper, we introduce LEGO-GraphRAG, a unified framework for the modular analysis and design of GraphRAG instances. By dividing the GraphRAG retrieval process into distinct modules and identifying corresponding design solutions, LEGO-GraphRAG provides a viable approach for building advanced GraphRAG systems.
Building upon the LEGO-GraphRAG framework, we conduct extensive empirical studies on large-scale real-world graphs and diverse GraphRAG query sets, leading to key findings:
{\bf 1)} GraphRAG must balance quality, efficiency, and cost across three aspects: modules, method types, and semantic models.
{\bf 2)} Extracting query-relevant subgraphs is the primary efficiency bottleneck for GraphRAG on large-scale graphs and remains underexplored.
{\bf 3)} Graph structural information enables efficient solutions for GraphRAG, while semantic information is crucial for improving the quality of complex queries. The optimal solution should integrate structural information to identify query-relevant subgraphs and leverage semantic information to retrieve reasoning paths.

%% file: section/acknowledgement.tex
\begin{acks}
This work was supported in part by the National Natural Science Foundation of China under Grant 62472400, Grant 62072428, Grant 62271465, in part by the Suzhou Basic Research Program under Grant SYG202338, and in part by the HK RGC grants 12202024, R1015-23, and C1043-24GF. Yukun Cao and Zengyi Gao contributed equally to this work. Xike Xie is the corresponding author.
\label{section: acknowledgements}
\end{acks}